\def\year{2021}\relax
\documentclass[letterpaper]{article} 
\usepackage{aaai21}  
\usepackage{times}  
\usepackage{helvet} 
\usepackage{courier}  
\usepackage[hyphens]{url}  
\usepackage{graphicx} 
\usepackage{natbib}  
\usepackage{caption} 
\usepackage{amsmath}
\usepackage{amssymb}
\title{Temporal-Coded Deep Spiking Neural Network with Easy Training and Robust Performance}

\author{
   Shibo Zhou,\textsuperscript{\rm 1} 
   Xiaohua Li,\textsuperscript{\rm 1}
   Ying Chen,\textsuperscript{\rm 2}
   Sanjeev T. Chandrasekaran,\textsuperscript{\rm 3}
   Arindam Sanyal \textsuperscript{\rm 3} \\
}
\affiliations{
\textsuperscript{\rm 1} Dept. of ECE, Binghamton University, Binghamton, NY, USA \\
\textsuperscript{\rm 2} (Corresponding Author) Harbin Institute of Technology, China \\
\textsuperscript{\rm 3} Dept. of EE, University at Buffalo, Buffalo, NY, USA \\
\{szhou19, xli\}@binghamton.edu, yingchen@hit.edu.cn, \{stannirk, arindams\}@buffalo.edu
}

\begin{document}

\maketitle

\begin{abstract}
Spiking neural network (SNN) is promising but the development has fallen far behind conventional deep neural networks (DNNs) because of difficult training. To resolve the training problem, we analyze the closed-form input-output response of spiking neurons and use the response expression to build abstract SNN models for training. This avoids calculating membrane potential during training and makes the direct training of SNN as efficient as DNN. We show that the nonleaky integrate-and-fire neuron with single-spike temporal-coding is the best choice for direct-train deep SNNs. We develop an energy-efficient phase-domain signal processing circuit for the neuron and propose a direct-train deep SNN framework. Thanks to easy training, we train deep SNNs under weight quantizations to study their robustness over low-cost neuromorphic hardware.  Experiments show that our direct-train deep SNNs have the highest CIFAR-10 classification accuracy among SNNs, achieve ImageNet classification accuracy within 1\% of the DNN of equivalent architecture, and are robust to weight quantization and noise perturbation.
\end{abstract}


\section{Introduction}

Spiking neural network (SNN) is interesting theoretically due to its strong bio-plausibility and practically because of its outstanding energy efficiency. Neurons communicate via spikes just as biological neurons. They work asynchronously, i.e., generate output spikes without waiting for all input neurons to spike. This leads to advantages such as spike sparsity, low latency, and high energy efficiency that are attractive for practical applications \cite{pfeiffer2018deep,tavanaei2019deep}.

The performance of SNNs has fallen far behind conventional deep neural networks (DNNs). One of the primary reasons is that SNNs are difficult to train. DNNs are formulated with the standard layer response ${\bf y}=f({\bf x}{\bf W} + b)$ where gradient backpropagation can be efficiently conducted. In contrast, for SNNs we have to simulate the temporal-domain neuron membrane potentials with non-differentiable spikes. Gradient evaluation is both difficult and time-consuming. Direct training of SNN has so far been limited to shallow networks only. No one has trained directly the SNNs on large datasets such as ImageNet. 

SNNs are targeting neuromorphic hardware implementations. Their robustness on low-cost neuromorphic hardware with limited memory, high noise, and large parameter drifting is critical for them to be competitive to DNNs in practical applications. This special yet important problem has been largely open.



In this paper, we resolve the difficult training problem by developing a direct-train framework that does not need calculating neuron membrane potentials. 
We address the robustness problem by developing a neuron circuit to evaluate timing jitter and by training deep SNNs under weight quantizations.
Our major contributions are listed as follows.

\begin{itemize}
    \item Closed-form analytical input-output responses of spiking neurons are studied systematically, which shows that the nonleaky integrate-and-fire neuron with single-spike temporal-coding is the best choice for direct-train SNNs.
    
    \item Deep SNNs such as SpikingVGG16 and SpikingGoogleNet are developed and trained over the CIFAR-10 and ImageNet datasets. New benchmark results are obtained. To the best of our knowledge, this is the first time that direct-train SNN is reported for ImageNet.
    
    \item A phase-domain signal processing neuron circuit is designed to show that our neuron is more energy-efficient than others and is robust to input timing jitter and weight quantization. Besides, our deep SNNs are trained under weight quantization and noise perturbation to demonstrate their robustness. 
\end{itemize}

This paper is organized as follows. Related works are introduced in Section \ref{sec2}. Spiking neurons and deep SNNs are described in Section \ref{sec3}. Experiments are presented in Section \ref{sec4}. Conclusions are given in Section \ref{sec5}.

\section{Related Works} \label{sec2}

SNN training methods can be categorized into three classes: unsupervised learning, supervised learning with indirect training, and supervised learning with direct training \cite{pfeiffer2018deep}. 
For unsupervised learning, spike timing-dependent plasticity (STDP) is well known 
\cite{caporale2008spike,diehl2015unsupervised,kheradpisheh2018stdp,lee2018deep}. Nevertheless, the dependency on the local neuronal activities without a global supervisor makes it have low performance in deep networks.

For the second class, the popular approach is to translate pre-trained DNNs to SNNs \cite{tavanaei2019deep}. This can be conducted by mapping DNN neuron values to SNN neuron spike rates \cite{diehl2015fast,rueckauer2017conversion,sengupta2019going} 
or spike times \cite{zhang2019tdsnn}. 
Although the translation approach has had the best performance among SNNs so far, it unfortunately sacrifices important SNN advantages such as spike sparsity and asynchronous processing, which degrades energy efficiency. Besides, many DNN techniques such as max-pooling and tanh activation are hard or inefficient to copy to SNN.


Among the third class, SpikeProp \cite{bohte2002error} minimized the loss between the true and desired spike times with gradient descent rule over soft nonlinearity models. Gardner et al. \cite{gardner2015learning} applied a probability neuron model to calculate gradients. Gradient backpropagation was applied in \cite{hunsberger2016training,lee2016training,jin2018hybrid,wu2019direct} based on spike rate coding where gradient backpropagation through both SNN layer responses and neuron membrane potential dynamics was needed, which made the algorithms extremely complex. Backpropagation through SNN layer responses only was applied in \cite{mostafa2017supervised}. Zhou and Li \cite{zhou2020spiking} pointed out that the overly-nonlinear neuron response led to low training performance in SNNs. 
Existing direct training approaches still fall short of efficiency to deal with large datasets such as ImageNet.

A list of neuromorphic hardware has been developed for SNN, such as IBM TrueNorth \cite{merolla2014million}, Intel Loihi \cite{davies2018loihi}, and BrainScaleS \cite{aamir2018accelerated}. 
For energy efficiency, Hunsberger and Eliasmith \cite{hunsberger2016training} estimated that a synaptic operation consumed only 8\% of the energy of a microprocessor floating-point operation. Cao et al. \cite{cao2015spiking} showed that SNN implemented in a neuromorphic circuit with 45 pJ per spike was 185 times more energy-efficient than the FPGA-based DNN implementation. With 26 pJ per spike, IBM TrueNorth consumed $1.76\times 10^5$ times less energy than computer simulations over microprocessors, and 769 times less energy than microprocessor-based neuromorphic hardware. Zhou et al. \cite{zhou2020deep} developed an analog neuron circuit with $19$ pJ per spike, with which SNN-based YOYOv2 consumed only 0.247 mJ for detecting objects in an image frame. 

Neuromorphic hardware has the problem of noise, parameter drifting, as well as severe limitation on resources such as memory. Implementing SNNs in neuromorphic hardware without robustness optimization showed heavy performance degradation \cite{esser1603convolutional,goltz2019fast}. Rathi et al. \cite{rathi2018stdp} studied the pruning of unimportant weights and quantizing important weights to improve energy efficiency, but for shallow SNNs only. 

\section{Deep SNN with Easy and Direct Training}
\label{sec3}



The major hurdle to training SNNs is that every neuron's membrane potential has to be calculated in each training iteration. This is not a problem when implementing SNNs in neuromorphic hardware but is computationally prohibitive for software implementation and gradient-based training. Non-differential spike waveform is the second hurdle. Our objective is to completely avoid the calculation of membrane potential and spike waveform during training. Instead, we train the SNNs based on an abstract {\it layer response model} of the neurons. As illustrated in Fig. \ref{fig:neuronfigure-append}, the left figure is the spiking neuron model for real hardware implementation and inference only, where the weights $w_{ji}$ need to be learned. The right figure is its layer response model, where we use the closed-form input-output response $t_j=f(t_i, w_{ji})$ to train the weights $w_{ji}$. Since the weights of the two figures are identical, the weights trained in the right figure are directly used to implement the SNN in the left figure. 

In this section, we first show that only a special spiking neuron is appropriate for such direct training. Then, we show that this spiking neuron can be practically realized in energy-efficient hardware. Finally, based on this neuron, we propose a direct-train framework for deep SNNs. 

\subsection{Layer Response Models of Spiking Neurons} \label{subsection31}

\begin{figure}[t]
\centering
\includegraphics[width=1.0\linewidth]{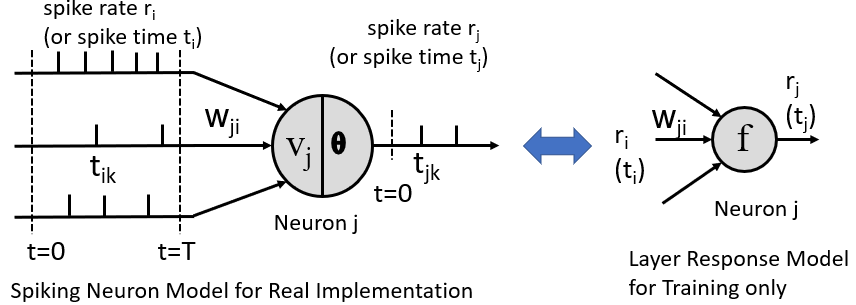}
\caption{Integrate-and-fire spiking neuron model.}
\label{fig:neuronfigure-append}
\end{figure}

For the neuron illustrated in Fig. \ref{fig:neuronfigure-append} (left), we consider the integrated-and-fire model. Membrane potential $v_j(t)$ of the neuron $j$ is modeled as
\begin{equation}
    \frac{dv_j(t)}{dt}+bv_j(t) = \sum_i w_{ji} \sum_{k} g(t-t_{ik}),  \label{eq2.10}
\end{equation}
where $b$ is a positive constant representing the leaky rate of the membrane potential, $w_{ji}$ is the weight of the synaptic connection from the input neuron $i$ to the output neuron $j$,  $g(t)$ is the synaptic current kernel function or spike waveform, and $t_{ik}$ is the spiking time of the $k$th spike of the $i$th input (pre-synaptic) neuron. $b>0$ means {\it leaky integrate-and-fire} (LIF) neuron, while $b=0$ means {\it nonleaky integrate-and-fire} (IF) neuron. 
Once $v_j(t)$ reaches spiking threshold $\theta$, the neuron generates an output (post-synaptic) spike and the membrane potential is reset. 

Information can be encoded in spike rate $r_j$, spike time $t_j$, or other means. We consider the first two, which we call rate coding and temporal coding, respectively. Rate $r_j$ is the average spike rate from $t=0$ to $t=T$. For temporal coding, each neuron generates a single spike during the time period $T$. We denote the spike time as $t_j$ and adopt the time-to-first-spike (TTFS) code \cite{goltz2019fast}. 

We have conducted a thorough study of the solutions to (\ref{eq2.10}) in order to find desirable layer response models. Details are in Technical Appendix \ref{secA.3}. We analyze our major observations only in this subsection.

Consider rate coding first. With impulse spike $g(t)=\delta(t)$, nonleaky IF neuron has closed-form layer response
\begin{equation}
    r_j = {\rm ReLU}\left(\sum_i r_i \frac{w_{ji}}{\theta} \right), \label{eq2.20}
\end{equation}
where ${\rm ReLU}(x)=\max\{0, x\}$. See (\ref{eq5.40}) for derivations. Similar expressions exist for Heaviside and exponentially-decaying spike waveforms (\ref{eq5.45})(\ref{eq5.50}). Since these expressions are identical to DNN's layer response, we can directly train a network implemented in software based on (\ref{eq2.20}) and apply the resulted weights $w_{ji}$ to the real SNN implemented in neuromorphic hardware. 

Note that (\ref{eq2.20}) is also the theoretical basis for translating DNNs to SNNs \cite{cao2015spiking, diehl2015fast}. It is interesting to see that direct-train SNN and translate-SNN become similar based on (\ref{eq2.20}). The only difference is that the latter trains weights $w_{ji}/\theta$ instead of $w_{ji}$ and thus needs weight normalization. Unfortunately, (\ref{eq2.20}) is an approximate model only. Modeling error accumulates to the detrimental level in deep SNNs \cite{rueckauer2017conversion}. Corrections are developed for translate-SNNs to mitigate the error to some extent. The corrections need to calculate membrane potential, which makes direct-train difficult again.

For rate-coded LIF neurons, layer responses become numerically unstable for training. As shown in (\ref{eq5.60}), LIF neuron with the impulse spike waveform has layer response
\begin{equation}
    r_j = {\rm ReLU}(-b \log^{-1}(1-\theta/\sum_i w_{ji}/(e^{b/r_i}-1))). \label{eq2.30}
\end{equation}
Random weights $w_{ji}$ often make $\log$ function undefined, which means training can not proceed. The same problem happens for other spike waveforms (\ref{eq5.65})(\ref{eq5.68}).  

Next, for temporal coding, 
similar numerical instability happens for LIF neurons with the exponentially-decaying spike waveform. As shown in (\ref{eq3.74}) and (\ref{eq3.75}), the layer responses are expressed in either Lambert W function or quadratic equation roots. Random weights often result in negative or complex values that prevent gradient updating.

Fortunately, temporal-coded nonleaky IF neurons have layer responses desirable for direct training. With the exponentially-decaying spike waveform, the layer response can be formulated as \cite{mostafa2017supervised}
\begin{equation}
    e^{\frac{t_j}{\tau}} = \sum_{i \in {\cal C}} e^{\frac{t_i}{\tau}} \frac{w_{ji}}{\sum_{\ell \in {\cal C}} w_{j\ell}-\theta}   \label{eq2.50}
\end{equation}
where the set ${\cal C}=\{\forall k: t_k < t_j\}$. See (\ref{eq3.26}) for derivations. With the Heaviside spike waveform, both IF and LIF neurons have similar layer responses, see (\ref{eq3.25}) and (\ref{eq3.65}). There is no significant modeling error and the expressions have nice numerical stability.

Since the energy efficiency of SNNs depends on the number of spikes, we prefer single-spike neurons. Therefore, the best choice is to adopt the single-spike temporal-coded nonleaky IF neuron to implement SNNs and train them based on (\ref{eq2.50}). Note that we prefer the exponentially-decaying spike waveform rather than the Heaviside spike waveform because the former can be truncated to short spike duration in practice to enhance energy efficiency. 


\subsection{Circuit of Temporal-Coded Spiking Neuron}

Most existing neuron circuits are designed for rate coding or for generating special spike patterns \cite{aamir2018mixed}. Few circuits are reported for single-spike neurons. Therefore, we present a new neuron circuit design to show that the single-spike temporal-coded neuron with exponentially-decaying spike waveform can be realized practically. This circuit will also help us study energy efficiency, timing jitter, and weight quantization robustness.

\begin{figure*}[t]
\centering
\includegraphics[width=0.85\textwidth]{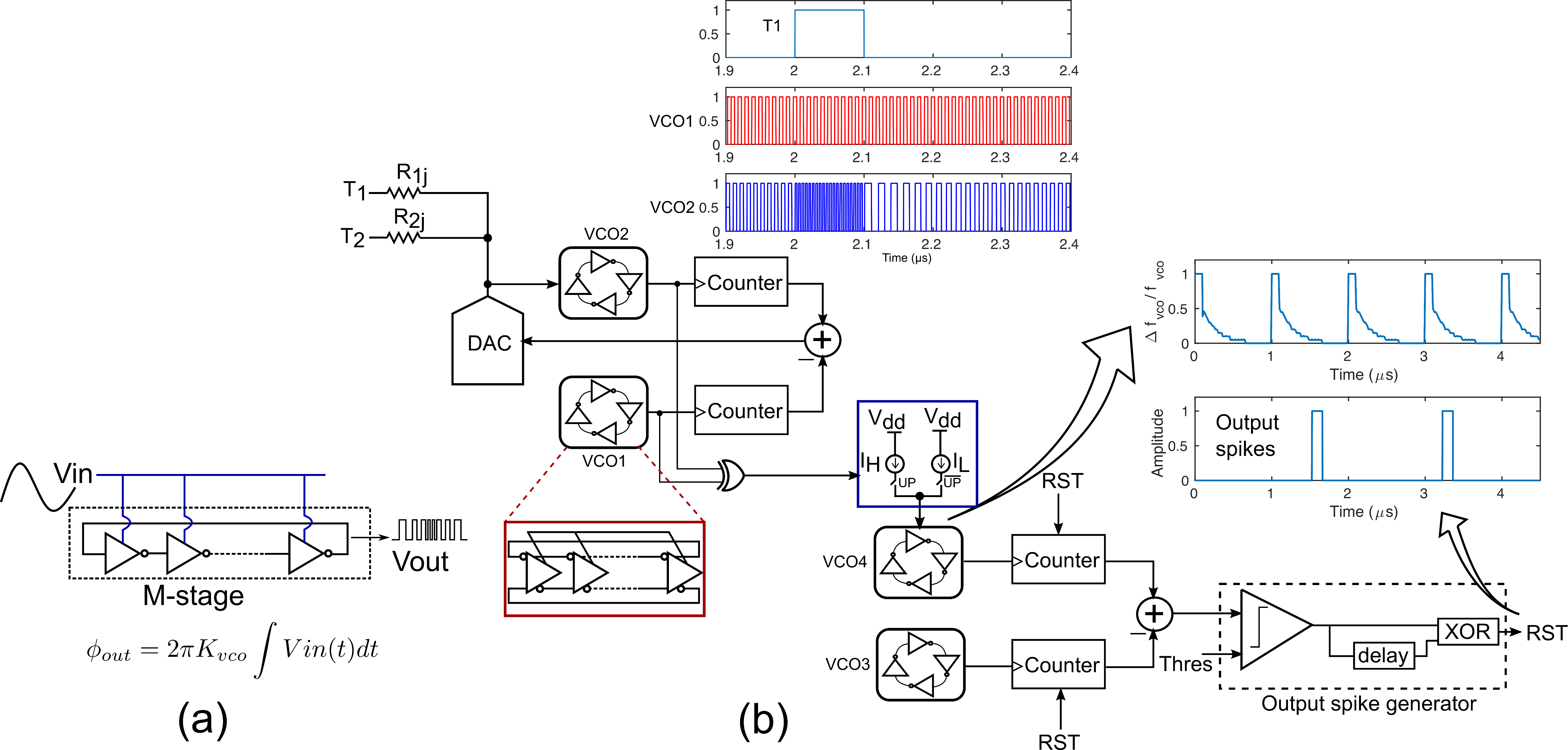}
\caption{IF neuron circuit schematic using time-domain signal processing circuits. (a) Operation of ring-inverter-based VCO. (b) Circuit schematic of 1 neuron designed using ring VCOs. Waveforms of input, output and intermediate nodes show that T$_1$ generates an input spike every microsecond, VCO1/VCO2 reshape the spike into the exponentially-decaying spike waveform, and VCO3/VCO4 generate an output spike every two microseconds. 
}
\label{fig:neuroncircuit}
\end{figure*}

For neuron circuit designs, prior works have used analog voltage domain (VD) signal processing to realize LIF neurons using CMOS circuits~\cite{indiveri2003low,wijekoon2009cmos,wu2015cmos}. VD signal processing techniques require high-gain amplifiers which are energy inefficient to design in advanced CMOS nodes with low intrinsic gain.
To reduce energy consumption, we introduce phase-domain (PD) signal processing into SNN design. PD signal processing converts VD signal excursions into phase domain for further processing, and has well documented advantages over VD processing such as higher energy efficiency, ability to better leverage CMOS technology scaling and simpler circuit design. While PD signal processing has been widely adopted in the circuit design community for high energy-efficiency data-converter design~\cite{sanyal2014hybrid,sanyal201618,jayaraj2019highly,akshay_asscc}, a key contribution of this work is the introduction of PD signal processing to SNN design field for the first time. 


Fig. \ref{fig:neuroncircuit} shows the circuit schematic of the IF neuron 
designed with voltage-controlled ring oscillator (VCO) based integrators. 
As shown in Fig.~\ref{fig:neuroncircuit}(a), if a voltage input $V_{in}(t)$ is applied to a ring VCO, its instantaneous phase is given by
$\Phi(t)=\int{2\pi k_{vco}  V_{in}(t)dt}$,
where $k_{vco}$ is VCO tuning gain. Thus, a VCO acts as a perfect PD integrator with infinite dc gain \cite{sanyal201618, taylor2010mostly, perrott200812}, and is perfectly positioned to realize non-leaky integration.  While the VCO phase is integral of its input, the phase output cannot be directly extracted. Rather, the voltage output of one of the inverters is read out which acts as a pulse-width modulated signal (PWM) and encodes the phase information in the width of its pulses as shown in Fig.~\ref{fig:neuroncircuit}(a).
The ring VCO has a highly digital architecture and can operate from very low supply voltages. Hence, VCO can be used as an integrator in advanced CMOS processes which allows area and power scaling unlike VD integrators. 

Fig.~\ref{fig:neuroncircuit}(b) shows the schematic of the proposed phase-domain IF neuron designed using VCOs, in which VCO1 and VCO3 act as global reference sources driven by constant inputs and are shared with all the neurons.
 $T_1$ and $T_2$ represent $2$ inputs to the neuron while the resistors $R_{1j}$ and $R_{2j}$ denote the weights of synaptic connections of the inputs to the neuron, respectively. When the inputs to VCO2 spike, the output phase of the VCO2 jumps abruptly and VCO2 starts running at a high frequency. Outputs of VCO1 and VCO2 drive two counters which increment at the rising edge of VCO outputs. The difference between the two counters is used to drive VCO2 through negative feedback.
 The negative feedback loop forces VCO2 to track VCO1, and the frequency of VCO2 starts decaying exponentially till it reaches the frequency of VCO1 as shown in Fig.~\ref{fig:neuroncircuit}(b). Thus, VCO2 realizes the exponential decaying synaptic kernel. The frequency difference between VCO1 and VCO2 is extracted using an XOR gate and sent to VCO4 for accumulation. VCO4 switches between two frequencies depending on whether the XOR output is high or low. Similar to VCO1 and VCO2, VCO3 and VCO4 also drive two counters at rising edges of their respective outputs.
  Once the difference between the two counter outputs exceeds a threshold value, an output spike is generated, and both the counters are reset.

The proposed circuit is highly scalable since it is built using digital CMOS circuits. The synaptic weights can be made tunable by using a digitally controllable resistor bank.
The neuron consumes less than $10$ pJ/spike in $65$nm process and the energy consumption will reduce further with technology scaling. This leads to an energy efficiency gain of $45\%$ over \cite{zhou2020deep}.  




\subsection{Direct-Train Framework for Deep SNNs}


For training, we implement the abstract SNNs in software based on (\ref{eq2.50}). Specifically, in the $\ell$th layer, $z_{\ell-1,i} = e^{t_{\ell-1,i}/\tau}$ and $z_{\ell,j} = e^{t_{\ell,j}/\tau}$ are used directly as neuron's input and output. For an $L$-layer deep SNN, define the input as ${\bf z}_0$ with elements $z_{0,i}$ and the final output as ${\bf z}_{L}$ with elements $z_{L, i}$. Smaller $z_{L, i}$ means stronger classification output. Then we have ${\bf z}_{L} = f({\bf z}_0; {\bf w})$ with nonlinear mapping $f$ and trainable weight ${\bf w}$ which includes all weights $w_{ji}^\ell$. Let the targeting output be class $c$. We train the network with the loss function
\begin{align}
    &{\cal L}  ({\bf z}_L,  c) = -\log \frac{z_{L, c}^{-1}}{\sum_{i\neq c} z_{L, i}^{-1}} + \nonumber \\
    &  K \sum_{\ell=1}^L \sum_j \max\left\{0, \theta- \sum_i w_{ji}^\ell \right\} + \lambda \sum_{\ell=1}^L \sum_{j, i} (w_{ji}^{\ell})^2.   \label{eq3.50}
\end{align}
The first term is to make $z_{L, c}$ the smallest (equivalently $t_{L, c}$ the smallest) one. The second term is the weight sum cost, which enlarges each neuron's input weight summation to increase its firing probability. The third term is $L_2$ regularization to prevent weights from becoming too large. The parameters $K$ and $\lambda$ are weighting coefficients. 
Thanks to the closed-form expression (\ref{eq2.50}), gradient backpropagation can be used to train the weights. The training becomes nothing different from conventional DNNs. 

Equations (\ref{eq2.50}) and (\ref{eq3.50}) were initially given in \cite{mostafa2017supervised}. However, only shallow networks with $2\sim 3$ fully-connected layers were tried and the performance was low. One of the problems of the algorithm presented in \cite{mostafa2017supervised} is that
 $t_j> t_i$ for $i \in {\cal C}$ was not checked, which led to $t_j \leq t_i$ or even negative timing that was not hardware realizable. 
Another problem is that the algorithm is not suitable for deep SNN. As a result, this work did not arouse too much interest and the research progress has been slow. 

We have resolved these problems by developing the following new algorithm to calculate (\ref{eq2.50}): 1) Use the SORT function to sort inputs $e^{t_i/\tau}$; 2) Use the CUMSUM function to list all possible $\sum_{i\in {\cal C}} w_{ji}$ and $\sum_{i \in {\cal C}} w_{ji} e^{t_i/\tau}$; 3) Calculate $e^{t_j/\tau}$ for each element in the list; 4) Find the first one that satisfies $e^{t_j/\tau} > e^{t_i/\tau}$ for $i\in {\cal C}$ and $e^{t_j/\tau} \leq e^{t_i/\tau}$ for $i \not\in {\cal C}$. Refer to our source code for details. This algorithm is much easier to implement and more efficient to run for deep SNNs. Note that real hardware SNN does not need this algorithm because spike time is naturally ordered.


As long as $\sum_i w_{ji}>\theta$, our algorithm can always provide hardware-compatible solutions $e^{t_j/\tau}$. With some heuristic convergence improving techniques (see experiment section), we have improved greatly both computation speed and convergence, which makes it possible to train deep SNNs. 



To design deep SNNs, it is helpful to follow DNN architectures so as to exploit the extensive DNN research experience. For this we need to study how some DNN-specific techniques can be realized in SNNs, such as pooling, batch normalization (BN), and local response normalization (LRN).
This issue has never been addressed in direct-train SNN including \cite{mostafa2017supervised}. 

For max-pooling, in software SNN training we have $e^{t_j/\tau} = \min_i e^{t_i/\tau}$. In real SNN hardware this can be realized as Fig. \ref{fig:neuronfigure-append} using a large enough $w_{ji}$, which lets the first arrival spike activate the output spike. Average-pooling can be realized by letting $w_{ji}=\theta/(N-1)$ for $N$ inputs. 

For BN, in software abstract SNN we need to calculate $e^{t_j/\tau}=\gamma/\sigma(e^{t_i/\tau}-\mu)+\beta$, where $\gamma, \sigma, \mu, \beta$ are parameters obtained from training. Rewriting it as
\begin{equation}
 e^{t_j/\tau} = e^{(t_i+\tau \log \gamma/\sigma)/\tau} + e^{\log(\beta-\gamma\mu/\sigma)}, \label{eq2.80}
\end{equation}
it is easy to see that (\ref{eq2.80}) can be realized in SNN hardware with two input neurons: one has spike time $t_i$ with a constant delay $\tau\log \gamma/\sigma$, and the other has a constant spike time $\tau\log(\beta-\gamma\mu/\sigma)$. The weight should be $w_{ji}=\theta$. Note that spike time delay is a standard feature in SNN hardware such as IBM TrueNorth. 

For LRN, the implementation is similar to BN but is much more complex. We also need to exploit the asynchronous property of SNN, i.e., reference timing $t=0$ of a layer can be shifted to some non-zero value. In software SNN we have $e^{t_j/\tau} = e^{t_i/\tau}/(\gamma + \alpha \sum_{\ell} e^{2t_\ell/\tau})^\beta$ for constants $\gamma$, $\alpha$ and $\beta$. In hardware, we can use a neuron to calculate  $\sum_{\ell}e^{2t_\ell/\tau}$ as average-pooling and then use another set of neurons to realize the rest operations similar to BN.  

The above DNN-adaptation techniques would make our direct-train SNN look similar to translate-SNN. This is not surprising as our abstract SNN is designed to be trained similarly as DNN. The major differences are, first, we directly train the real SNN weights $w_{ji}$ and thus do not need complex weight normalization or SNN structure change. Second, we have the flexibility to improve DNN techniques and train them to fit with SNN. For example, in BN, we can train SNN to guarantee the output in (\ref{eq2.80}) be non-negative. Third, we can develop SNN specific techniques to outperform DNNs, such as adjusting $\theta$ in (\ref{eq3.50}) to promote sparsity and energy efficiency. Note that besides spike sparsity, our temporal-coded SNN has neuron sparsity, i.e., only neurons in ${\cal C}$ need to spike. Translate-SNN does not have such sparsity. The sparsity of translate-SNN usually means to reduce the number of spikes by using a shorter $T$, which makes the SNN work in transient response mode and thus suffer from performance degradation.

\section{Experiments}
\label{sec4}

In this section, we report our experiments over three standard datasets: MNIST, CIFAR-10, and ImageNet\footnote{Our source code can be found at \url{https://github.com/zbs881314/Temporal-Coded-Deep-SNN}}. 
We designed three deep SNN models, which are shown in Table \ref{tbl:modellist}, and evaluated their accuracy and robustness.   

\begin{table*}[t]
  \centering
  \begin{tabular}{c|l}
    \hline
   MNIST & SCNN(5,32,2) $\rightarrow$ SCNN(5,16,2) $\rightarrow$ FC(10) \\
   CIFAR-10 & {\bf SpikingVGG16}: SCNN(3,64,1) $\rightarrow$ SCNN(3,64,1) $\rightarrow$ MP(2) $\rightarrow$ SCNN(3,128,1) 
        $\rightarrow$  SCNN(3,128,1)  \\ 
        & $\rightarrow$ MP(2)
         $\rightarrow$ SCNN(3,256,1) $\rightarrow$ SCNN(3,256,1) 
      $\rightarrow$  SCNN(3,256,1) $\rightarrow$  MP(2) $\rightarrow$ SCNN(3,512,1) \\ 
      & $\rightarrow$ SCNN(3,512,1) 
     $\rightarrow$ SCNN(3,512,1) $\rightarrow$ MP(2) $\rightarrow$ SCNN(3,1024,1) $\rightarrow$ SCNN(3,1024,1)  \\
    & $\rightarrow$ SCNN(3,1024,1) $\rightarrow$ MP(2) 
     $\rightarrow$ FC(4096) $\rightarrow$ FC(4096) $\rightarrow$ FC(512) $\rightarrow$ FC(10) \\
   ImageNet & {\bf SpikingGoogleNet}: replace GoogleNet CNN/FC layers with SCNN/FC layers \\
    \hline
  \end{tabular}
    \caption{Proposed models. SCNN(5,32,2) means spiking CNN layer with 32 $5\times5$ kernels and stride 2. FC(10) means fully-connected layer with 10 output neurons. MP(2) means $2\times2$ max-pooling layer. $\tau=\theta=1$. Bias is added as $t_0=0$.}
  \label{tbl:modellist}
\end{table*}

\subsection{Classification Accuracy} \label{sec41}

\paragraph{MNIST:} The MNIST image pixels were normalized to $p_i \in [0, 1]$ and encoded into spiking time $t_i=\alpha(-p_i+1)$. The parameter $\alpha$ was used to adjust spike temporal separation. We trained the network for 50 epochs with batch size $10$ using the Adam optimizer. The learning rate started at $0.001$ and gradually reduced to $0.0001$ at the last epoch, with learning decay lr\_decay = (learning\_end -learning\_start)/50.  We set $K=100$ and $\lambda=0.001$.  According to (\ref{eq2.50}), gradients could become very large in case $\sum_{\ell}w_{j\ell}$ is near $\theta$, which is harmful to training. Therefore, we limited the maximum allowed row-normalized Frobenius norm of the gradient of each weight matrix to $10$.   

We trained this network with noisy input spike times. Classification accuracy is shown in Table \ref{tbl:mnistacc}. The proposed network had the highest accuracy (99.33\%) among the SNNs listed in the table, yet had the smallest network size with the least number of trainable weights (21K). Because of the asynchronous operation of our SNN, on average $94\%$ neurons spiked, which led to sparsity 0.94. The total consumed energy of the proposed network was $205$ nJ assuming each spike cost $10$ pJ based on the proposed neuron circuit and sparsity 0.94. 

\begin{table*}[t]
  \centering
  \begin{tabular}{c|ccccc}
    \hline
Dataset &  Models  &  Method & Accuracy & Weights(million) & Sparsity  \\
    \hline
   & \cite{ciresan2011convolutional} & CNN & 99.73\%  & 0.069 & no \\
 &   
    \cite{kheradpisheh2018stdp} & SNN+STDP	& 98.40\%	& 0.076 & no \\
  & \cite{lee2018deep} &	SNN+STDP	& 91.10\%	& 0.025 & no \\
MNIST  & \cite{mostafa2017supervised} &	SNN+DT &	97.55\%	& 0.635 & 0.51 \\
  & \cite{zhang2019tdsnn} & SNN+tran &	99.08\%	 & 3.9  &  no \\
  &  \cite{wu2019deep} &	SNN+DT &	99.26\%	& 0.051  & no \\
  & {\bf Our Model} &	{\bf SNN+DT} &	{\bf 99.33\%}	& {\bf 0.021} & {\bf 0.94} \\
\hline \hline

  &  \cite{liu2015very} (VGG16) & CNN & 91.55\%  &  15  & no \\
&  \cite{hunsberger2016training} & SNN+tran	& 82.95\%	& 39 & no \\
CIFAR-10  & \cite{rueckauer2017conversion} &	SNN+tran	& 90.85\%	& 62 & no  \\
  & \cite{sengupta2019going} (VGG16) & SNN+tran & 91.55\% & 15+  & no \\
  & \cite{wu2019direct} &	SNN+DT	& 90.53\% &	45  & no  \\
  & {\bf Our Model}	& {\bf SNN+DT}	& {\bf 92.68\%}	& {\bf 54}  & {\bf 0.62} \\

\hline \hline
& \cite{simonyan2014very} (VGG16) & CNN & 71.5\%   & 138  & no \\
& \cite{szegedy2015going} (GoogleNet)  & CNN & 69.8\%   & 6.8  & no \\
ImageNet & \cite{rueckauer2017conversion} (VGG16) & SNN+tran & 49.61\% & 138 & no \\
& \cite{sengupta2019going} (VGG16) & SNN+tran & 69.96\% & 138+ & no \\
&\cite{zhang2019tdsnn} (VGG16) & SNN+tran & 70.87\%   & $138+$ & no \\
   & {\bf Our Model}   & {\bf SNN+DT} &  {\bf 68.8\%}   & {\bf 6.8} & {\bf 0.56}   \\
    \hline
  \end{tabular}
  \caption{Classification accuracy comparison. DT: direct training. tran: translate SNN.}
  \label{tbl:mnistacc}
\end{table*}

\paragraph{CIFAR-10:} 
Our SNN model for CIFAR-10 was developed based on the VGG16 model \cite{liu2015very,simonyan2014very}, hence called SpikingVGG16. To encode image pixels to spike time, we applied encoding rule $t_i= \alpha p_i$ and used a relatively large $\alpha$ to enlarge spike time separation so that all pixel values could potentially be used. We exploited data augmentation such as crop, flip, and whiten to increase the diversity of data available for training.
The learning rate started at 0.01 and ended at 0.0001. We ran 320 epochs and after 240 epochs the training tended to converge. The batch size was 128. 
The other hyper-parameters were the same as the MNIST experiments. 

As shown in Table \ref{tbl:mnistacc}, the testing accuracy of our SpikingVGG16 set a new record at 92.68\%, higher than all the listed SNNs including the state-of-the-art of \cite{wu2019direct}. Especially, our model had higher accuracy than the CNN-based VGG16 \cite{liu2015very}. Our network enjoyed a sparsity of $0.62$ which means only $62\%$ neurons sent spikes. It consumed $0.336$ mJ of energy for each image inference. 


\paragraph{ImageNet:} We built our deep SNN model based on the popular GoogleNet architecture \cite{szegedy2015going}, hence called SpikingGoogleNet. We simply replaced CNN layers with SCNN layers and FC layers with spiking FC layers. We used 1.2 million images to train the network.
Training parameters were similar to the CIFAR-10 experiments, and the training procedure followed that of GoogleNet. The input image was $224 \times 224 \times 3$ and was randomly cropped from a resized image using the scale and aspect ratio augmentation. We used SGD with a batch size of 256, 
learning rate decay 0.0001, and momentum 0.9. We started from a learning rate of 0.1 and divided it by 10 three times. 
When the training error was no longer reducing,  retraining and fine-tuning with very small learning rates were conducted until the test accuracy no longer increased. 

Results of Top-1 testing accuracy are shown in Table \ref{tbl:mnistacc}. Our SpikingGoogleNet achieved 68.8\% accuracy, only 1.0\% lower than the CNN-based GoogleNet. VGG16-based SNNs had slightly better performance because CNN-based VGG16 had better accuracy than GoogleNet. But they had much larger network sizes.

For energy consumption, our network enjoyed a sparsity of $0.56$ which means only $56\%$ of neurons were activated during image inference. The total energy consumption was thus $0.038$ mJ.  For the translate-SNN model \cite{sengupta2019going}, if implemented in IBM TrueNorth with 26 pJ per spike and having an average of 256 spikes per neuron with rate coding, each of them would consume 900 mJ, several orders bigger than ours. 

To the best of our knowledge, few works were reported over the CIFAR-10 dataset using direct training SNN (except \cite{wu2019direct}) and none was reported over the ImageNet dataset. Our models hence set up new benchmark accuracy in both cases. There is still a big gap between SNN and DNN \cite{tan2019efficientnet}. Nevertheless, our experiments showed that our proposed SNNs could achieve similar accuracy as the DNNs with similar network size and architecture. This fact was also demonstrated in many translate SNN works. We expect that SNN performance could catch up rapidly after the training hurdle is resolved.

\subsection{Robustness} \label{sec42}

\paragraph{Neuron Robustness:} For neuromorphic circuits, a problem is that they are very sensitive to changes in operating conditions such as supply voltage or temperature. In Fig. \ref{fig:neuroncircuit}, we have used differential architecture in which VCO phase is always compared with phase from a reference VCO rather than using the absolute phase from a single VCO. While this architectural choice doubles area and energy consumption, it also increases robustness to changes in operating conditions. 

For temporal-coded neurons, an important distortion measure is spike timing jitter, which is defined as $|t_j^{\rm measured}-t_j^{\rm desired}|/t_j^{\rm desired}$. For mixed digital-analog circuits, there are quantization noise and circuit noise, which we simply combine into synaptic weight quantization noise. We conducted circuit simulation under various input timing jitters and weight quantizations, and the results are summarized in Table~\ref{tbl:vco}. Both input jitter and quantization of synaptic weights change the temporal location of the output spikes. As the input jitter was swept from 1\% to 9\%, the output jitter varied from 0.024\% to 0.997\% only. Such a good inherent jitter suppression capability was due to the phase-locked loop (PLL) which low-pass-filtered input jitter. 

Quantizing synaptic weights (for 2-input case) increased output jitter from 0.004\% in the case of 32-bit quantization to 0.957\% in the case of 4-bit quantization. Coarse quantization thus led to large jitter. One of the reasons was that in our neuron model (\ref{eq2.50}) the denominator $\sum_{\ell} w_{j\ell} -\theta$ could be small and a slight change of weights $w_{j\ell}$ might cause a big change in $t_j$. To mitigate this problem, we can set $\theta$ in (\ref{eq3.50}) to a bigger number during training. 

Examining the numbers in Table \ref{tbl:vco} more carefully, we see that 4-bit quantization, which has quantization SNR (signal to noise ratio) $6.02\times 4 \approx 24$ dB, led to output jitter of 0.957\% which means SNR $20\log_{10}(1/.00957)\approx 40$ dB. A $40$ dB SNR means that the neuron was extremely robust to weight quantization.

\begin{table}[t]
\centering
\begin{tabular}{|l|c|c|l|c|}
\cline{1-2} \cline{4-5}
Input  & Output  &  & Quantization & Output  \\
Jitter (\%) & Jitter (\%) & & Level & Jitter (\%) \\
\cline{1-2} \cline{4-5} 
1                 & 0.024              &  & 4-bit              & 0.957              \\ \cline{1-2} \cline{4-5} 
3                 & 0.189              &  & 8-bit              & 0.268              \\ \cline{1-2} \cline{4-5} 
5                 & 0.397              &  & 16-bit             & 0.008              \\ \cline{1-2} \cline{4-5} 
7                 & 0.789              &  & 24-bit             & 0.004              \\ \cline{1-2} \cline{4-5} 
9                 & 0.997              &  & 32-bit             & 0.004              \\ \cline{1-2} \cline{4-5} 
\end{tabular}
\caption{Jitter in output spike versus input jitter and synaptic weight quantization.}
\label{tbl:vco}
\end{table}

\paragraph{Deep SNN Robustness:} To evaluate the robustness of deep SNNs, we experimented with both weight quantization and noise perturbation. For weight quantization, the weights were quantized to 32-bit, 8-bit, 4-bit, and 2-bit words. Thanks to our easy training models, we could retrain the deep SNNs simply following the procedure developed for conventional CNN quantization \cite{li2016ternary,rastegari2016xnor}. Specifically, the forward inference used quantized weights while the backward gradient propagation used full-precision weights. We first trained with 32-bit quantization. After the training converged, we applied 8-bit quantization and retrain the SNNs. This procedure was repeated until the 2-bit quantization.

Table \ref{tbl:mnistaccquant} shows the testing accuracy under weight quantization. For the models over MNIST, CIFAR10 and ImageNet, weight quantization caused the worst accuracy loss of $0.22\%$, $1.75\%$, and $8.8\%$, respectively. As comparison, we listed two typical CNN weight quantization results \cite{cheng2018differentiable, zhang2018lq}, which indicated a similar performance degradation pace. The results demonstrated that the SNN models were robust to weight quantization for relatively small datasets and small networks. For larger networks, the weights would better be encoded in 4-bit or over. 

\begin{figure}[t]
\centering
\includegraphics[width=0.7\linewidth]{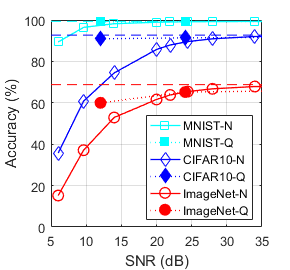}
\captionof{figure}{Accuracy versus weight quantization (Q) and noise perturbation (N). Dashed Lines: baseline (32-bit quantization and noiseless cases). Note: 8,4,2-bit quantizations correspond to quantization SNRs 48, 24, 12dB, respectively.}
\label{fig:noiseperturbation}
\end{figure}

To evaluate noise perturbation, we added random noise to the trained weights. Experiment results are shown in Fig. \ref{fig:noiseperturbation}, together with the results of weight quantization in Table \ref{tbl:mnistaccquant} (expressed in SNR). We find that 24dB quantization noise (4-bit quantization) reduced ImageNet classification accuracy to 65.2\%. Noise at 24dB SNR reduced ImageNet classification accuracy to 65.43\%. Both cases had a small 3.5\% performance loss only. 

Because Table \ref{tbl:vco} indicated that neuron noise (explained as jitter) was usually very small, we just need to pay attention to high SNR scenarios, such as 24dB and above (corresponding to 4-bit or above quantization). In this case, the accuracy reduction was negligible. Therefore, the deep SNNs were robust to both weight quantization and noise perturbation. 

Extra experiment results are in Technical Appendix \ref{appendixB}.

\begin{table}[t]
  \centering
  \begin{tabular}{ccccc}
    \hline
Model &   Bits     & MNIST  & CIFAR-10 & ImageNet    \\  
    \hline
 &    32 &  99.33\%  &  92.68\%  &    68.8\%   \\
Our &   8 &   99.32\%  &  91.87\%  &    66.1\%    \\
Models &   4 & 99.21\%  &  91.38\%   &    65.2\%  \\
 &   2&  99.11\%  &   90.93\%   &    60.0\%  \\  
 
 \hline
(Cheng et al.  &  32 &  98.66\% & 84.80\%  & -  \\
 2018) &  8 &  98.48\%  & 84.07\% & -  \\
  &  2  &  96.34\% &  81.56\% &  -  \\ \hline

(Zhang et al.  & 32 &- & 92.1\% & 70.3\% \\
2018) & 4 & - & - & 70.0\%  \\
& 2 &-  & 91.8\% &  68.0\% \\  
%

 \hline
  \end{tabular}
\caption{Accuracy versus weight quantization.}
  \label{tbl:mnistaccquant}
\end{table}

%
        
\section{Conclusion}
\label{sec5}

In this paper, we develop direct-train deep SNNs that can be easily trained over large datasets such as ImageNet. We show that our SNNs can realize classification accuracy within 1\% of (or even better than) DNNs of similar size and architecture. A circuit schematic of the adopted neuron is designed with phase-domain signal processing, which shows 45\% energy efficiency gain over the existing state of the art. Both the neuron and the deep SNNs are demonstrated as robust to timing jitter, weight quantization, and noise. The easy training, high performance, and robustness indicate that SNNs can be competitive to DNNs in practical applications.






\bibliography{aaai21}

\begin{thebibliography}{46}
\providecommand{\natexlab}[1]{#1}
\providecommand{\url}[1]{\texttt{#1}}
\providecommand{\urlprefix}{URL }
\expandafter\ifx\csname urlstyle\endcsname\relax
  \providecommand{\doi}[1]{doi:\discretionary{}{}{}#1}\else
  \providecommand{\doi}{doi:\discretionary{}{}{}\begingroup
  \urlstyle{rm}\Url}\fi

\bibitem[{Aamir et~al.(2018{\natexlab{a}})Aamir, M{\"u}ller, Kiene, Kriener,
  Stradmann, Gr{\"u}bl, Schemmel, and Meier}]{aamir2018mixed}
Aamir, S.~A.; M{\"u}ller, P.; Kiene, G.; Kriener, L.; Stradmann, Y.; Gr{\"u}bl,
  A.; Schemmel, J.; and Meier, K. 2018{\natexlab{a}}.
\newblock A mixed-signal structured adex neuron for accelerated neuromorphic
  cores.
\newblock \emph{IEEE transactions on biomedical circuits and systems} 12(5):
  1027--1037.

\bibitem[{Aamir et~al.(2018{\natexlab{b}})Aamir, Stradmann, M{\"u}ller, Pehle,
  Hartel, Gr{\"u}bl, Schemmel, and Meier}]{aamir2018accelerated}
Aamir, S.~A.; Stradmann, Y.; M{\"u}ller, P.; Pehle, C.; Hartel, A.; Gr{\"u}bl,
  A.; Schemmel, J.; and Meier, K. 2018{\natexlab{b}}.
\newblock An accelerated lif neuronal network array for a large-scale
  mixed-signal neuromorphic architecture.
\newblock \emph{IEEE Transactions on Circuits and Systems I: Regular Papers}
  65(12): 4299--4312.

\bibitem[{Bohte, Kok, and La~Poutre(2002)}]{bohte2002error}
Bohte, S.~M.; Kok, J.~N.; and La~Poutre, H. 2002.
\newblock Error-backpropagation in temporally encoded networks of spiking
  neurons.
\newblock \emph{Neurocomputing} 48(1-4): 17--37.

\bibitem[{Cao, Chen, and Khosla(2015)}]{cao2015spiking}
Cao, Y.; Chen, Y.; and Khosla, D. 2015.
\newblock Spiking deep convolutional neural networks for energy-efficient
  object recognition.
\newblock \emph{International Journal of Computer Vision} 113(1): 54--66.

\bibitem[{Caporale and Dan(2008)}]{caporale2008spike}
Caporale, N.; and Dan, Y. 2008.
\newblock Spike timing--dependent plasticity: a Hebbian learning rule.
\newblock \emph{Annu. Rev. Neurosci.} 31: 25--46.

\bibitem[{Cheng et~al.(2018)Cheng, Huang, Guo, Huang, Yan, Li, and
  Chen}]{cheng2018differentiable}
Cheng, H.-P.; Huang, Y.; Guo, X.; Huang, Y.; Yan, F.; Li, H.; and Chen, Y.
  2018.
\newblock Differentiable fine-grained quantization for deep neural network
  compression.
\newblock \emph{arXiv preprint arXiv:1810.10351} .

\bibitem[{Ciresan et~al.(2011)Ciresan, Meier, Gambardella, and
  Schmidhuber}]{ciresan2011convolutional}
Ciresan, D.~C.; Meier, U.; Gambardella, L.~M.; and Schmidhuber, J. 2011.
\newblock Convolutional neural network committees for handwritten character
  classification.
\newblock In \emph{2011 International Conference on Document Analysis and
  Recognition}, 1135--1139. IEEE.

\bibitem[{Davies et~al.(2018)Davies, Srinivasa, Lin, Chinya, Cao, Choday,
  Dimou, Joshi, Imam, Jain et~al.}]{davies2018loihi}
Davies, M.; Srinivasa, N.; Lin, T.-H.; Chinya, G.; Cao, Y.; Choday, S.~H.;
  Dimou, G.; Joshi, P.; Imam, N.; Jain, S.; et~al. 2018.
\newblock Loihi: A neuromorphic manycore processor with on-chip learning.
\newblock \emph{IEEE Micro} 38(1): 82--99.

\bibitem[{Diehl and Cook(2015)}]{diehl2015unsupervised}
Diehl, P.~U.; and Cook, M. 2015.
\newblock Unsupervised learning of digit recognition using
  spike-timing-dependent plasticity.
\newblock \emph{Frontiers in computational neuroscience} 9: 99.

\bibitem[{Diehl et~al.(2015)Diehl, Neil, Binas, Cook, Liu, and
  Pfeiffer}]{diehl2015fast}
Diehl, P.~U.; Neil, D.; Binas, J.; Cook, M.; Liu, S.-C.; and Pfeiffer, M. 2015.
\newblock Fast-classifying, high-accuracy spiking deep networks through weight
  and threshold balancing.
\newblock In \emph{2015 International Joint Conference on Neural Networks
  (IJCNN)}, 1--8. ieee.

\bibitem[{Esser et~al.(2016)Esser, Merolla, Arthur, Cassidy, Appuswamy,
  Andreopoulos, Berg, McKinstry, Melano, Barch et~al.}]{esser1603convolutional}
Esser, S.; Merolla, P.; Arthur, J.; Cassidy, A.; Appuswamy, R.; Andreopoulos,
  A.; Berg, D.; McKinstry, J.; Melano, T.; Barch, D.; et~al. 2016.
\newblock Convolutional Networks for Fast, Energy-E cient Neuromorphic
  Computing.
\newblock \emph{CoRR abs/.().: http://arxiv. org/abs/1603.08270} .

\bibitem[{Gardner, Sporea, and Gr{\"u}ning(2015)}]{gardner2015learning}
Gardner, B.; Sporea, I.; and Gr{\"u}ning, A. 2015.
\newblock Learning spatiotemporally encoded pattern transformations in
  structured spiking neural networks.
\newblock \emph{Neural computation} 27(12): 2548--2586.

\bibitem[{G{\"o}ltz et~al.(2019)G{\"o}ltz, Baumbach, Billaudelle, Breitwieser,
  Dold, Kriener, Kungl, Senn, Schemmel, Meier et~al.}]{goltz2019fast}
G{\"o}ltz, J.; Baumbach, A.; Billaudelle, S.; Breitwieser, O.; Dold, D.;
  Kriener, L.; Kungl, A.~F.; Senn, W.; Schemmel, J.; Meier, K.; et~al. 2019.
\newblock Fast and deep neuromorphic learning with time-to-first-spike coding.
\newblock \emph{arXiv preprint arXiv:1912.11443} .

\bibitem[{Hunsberger and Eliasmith(2016)}]{hunsberger2016training}
Hunsberger, E.; and Eliasmith, C. 2016.
\newblock Training spiking deep networks for neuromorphic hardware.
\newblock \emph{arXiv preprint arXiv:1611.05141} .

\bibitem[{Indiveri(2003)}]{indiveri2003low}
Indiveri, G. 2003.
\newblock {A low-power adaptive integrate-and-fire neuron circuit}.
\newblock In \emph{IEEE International Symposium on Circuits and Systems},
  volume~4, IV--IV.

\bibitem[{Jayaraj et~al.(2019{\natexlab{a}})Jayaraj, Danesh, Chandrasekaran,
  and Sanyal}]{jayaraj2019highly}
Jayaraj, A.; Danesh, M.; Chandrasekaran, S.~T.; and Sanyal, A.
  2019{\natexlab{a}}.
\newblock {Highly Digital Second-Order $\Delta\Sigma$ VCO ADC}.
\newblock \emph{IEEE Transactions on Circuits and Systems I: Regular Papers}
  66(7): 2415--2425.

\bibitem[{Jayaraj et~al.(2019{\natexlab{b}})Jayaraj, Das, Arcot, and
  Sanyal}]{akshay_asscc}
Jayaraj, A.; Das, A.; Arcot, S.; and Sanyal, A. 2019{\natexlab{b}}.
\newblock { 8.6fJ/step VCO-Based CT 2nd-Order $\Delta\Sigma$ ADC}.
\newblock In \emph{IEEE Asian Solid-State Circuits Conference (A-SSCC)},
  197--200.

\bibitem[{Jin, Zhang, and Li(2018)}]{jin2018hybrid}
Jin, Y.; Zhang, W.; and Li, P. 2018.
\newblock Hybrid macro/micro level backpropagation for training deep spiking
  neural networks.
\newblock In \emph{Advances in neural information processing systems},
  7005--7015.

\bibitem[{Kheradpisheh et~al.(2018)Kheradpisheh, Ganjtabesh, Thorpe, and
  Masquelier}]{kheradpisheh2018stdp}
Kheradpisheh, S.~R.; Ganjtabesh, M.; Thorpe, S.~J.; and Masquelier, T. 2018.
\newblock STDP-based spiking deep convolutional neural networks for object
  recognition.
\newblock \emph{Neural Networks} 99: 56--67.

\bibitem[{Lee et~al.(2018)Lee, Srinivasan, Panda, and Roy}]{lee2018deep}
Lee, C.; Srinivasan, G.; Panda, P.; and Roy, K. 2018.
\newblock Deep spiking convolutional neural network trained with unsupervised
  spike-timing-dependent plasticity.
\newblock \emph{IEEE Transactions on Cognitive and Developmental Systems}
  11(3): 384--394.

\bibitem[{Lee, Delbruck, and Pfeiffer(2016)}]{lee2016training}
Lee, J.~H.; Delbruck, T.; and Pfeiffer, M. 2016.
\newblock Training deep spiking neural networks using backpropagation.
\newblock \emph{Frontiers in neuroscience} 10: 508.

\bibitem[{Li, Zhang, and Liu(2016)}]{li2016ternary}
Li, F.; Zhang, B.; and Liu, B. 2016.
\newblock Ternary weight networks.
\newblock \emph{arXiv preprint arXiv:1605.04711} .

\bibitem[{Liu and Deng(2015)}]{liu2015very}
Liu, S.; and Deng, W. 2015.
\newblock Very deep convolutional neural network based image classification
  using small training sample size.
\newblock In \emph{2015 3rd IAPR Asian conference on pattern recognition
  (ACPR)}, 730--734. IEEE.

\bibitem[{Merolla et~al.(2014)Merolla, Arthur, Alvarez-Icaza, Cassidy, Sawada,
  Akopyan, Jackson, Imam, Guo, Nakamura et~al.}]{merolla2014million}
Merolla, P.~A.; Arthur, J.~V.; Alvarez-Icaza, R.; Cassidy, A.~S.; Sawada, J.;
  Akopyan, F.; Jackson, B.~L.; Imam, N.; Guo, C.; Nakamura, Y.; et~al. 2014.
\newblock A million spiking-neuron integrated circuit with a scalable
  communication network and interface.
\newblock \emph{Science} 345(6197): 668--673.

\bibitem[{Mostafa(2017)}]{mostafa2017supervised}
Mostafa, H. 2017.
\newblock Supervised learning based on temporal coding in spiking neural
  networks.
\newblock \emph{IEEE transactions on neural networks and learning systems}
  29(7): 3227--3235.

\bibitem[{Perrott et~al.(2008)}]{perrott200812}
Perrott, M.; et~al. 2008.
\newblock A 12-bit 10-MHz bandwidth continuous-time ADC with a 5-bit 950-MS/s
  VCO-based quantizer.
\newblock \emph{IEEE J. Solid-State Circuits} .

\bibitem[{Pfeiffer and Pfeil(2018)}]{pfeiffer2018deep}
Pfeiffer, M.; and Pfeil, T. 2018.
\newblock Deep learning with spiking neurons: opportunities and challenges.
\newblock \emph{Frontiers in neuroscience} 12: 774.

\bibitem[{Rastegari et~al.(2016)Rastegari, Ordonez, Redmon, and
  Farhadi}]{rastegari2016xnor}
Rastegari, M.; Ordonez, V.; Redmon, J.; and Farhadi, A. 2016.
\newblock Xnor-net: Imagenet classification using binary convolutional neural
  networks.
\newblock In \emph{European conference on computer vision}, 525--542. Springer.

\bibitem[{Rathi, Panda, and Roy(2018)}]{rathi2018stdp}
Rathi, N.; Panda, P.; and Roy, K. 2018.
\newblock STDP-based pruning of connections and weight quantization in spiking
  neural networks for energy-efficient recognition.
\newblock \emph{IEEE Transactions on Computer-Aided Design of Integrated
  Circuits and Systems} 38(4): 668--677.

\bibitem[{Rueckauer et~al.(2017)Rueckauer, Lungu, Hu, Pfeiffer, and
  Liu}]{rueckauer2017conversion}
Rueckauer, B.; Lungu, I.-A.; Hu, Y.; Pfeiffer, M.; and Liu, S.-C. 2017.
\newblock Conversion of continuous-valued deep networks to efficient
  event-driven networks for image classification.
\newblock \emph{Frontiers in neuroscience} 11: 682.

\bibitem[{Sanyal et~al.(2014)Sanyal, Ragab, Chen, Viswanathan, Yan, and
  Sun}]{sanyal2014hybrid}
Sanyal, A.; Ragab, K.; Chen, L.; Viswanathan, T.; Yan, S.; and Sun, N. 2014.
\newblock {A hybrid SAR-VCO $\Delta$$\Sigma$ ADC with first-order noise
  shaping}.
\newblock In \emph{IEEE Custom Integrated Circuits Conference}, 1--4.

\bibitem[{Sanyal and Sun(2016)}]{sanyal201618}
Sanyal, A.; and Sun, N. 2016.
\newblock A 18.5-fJ/step VCO-based 0--1 MASH $\Delta-\Sigma$ ADC with digital
  background calibration.
\newblock In \emph{2016 IEEE Symposium on VLSI Circuits (VLSI-Circuits)}, 1--2.
  IEEE.

\bibitem[{Sengupta et~al.(2019)Sengupta, Ye, Wang, Liu, and
  Roy}]{sengupta2019going}
Sengupta, A.; Ye, Y.; Wang, R.; Liu, C.; and Roy, K. 2019.
\newblock Going deeper in spiking neural networks: Vgg and residual
  architectures.
\newblock \emph{Frontiers in neuroscience} 13: 95.

\bibitem[{Simonyan and Zisserman(2014)}]{simonyan2014very}
Simonyan, K.; and Zisserman, A. 2014.
\newblock Very deep convolutional networks for large-scale image recognition.
\newblock \emph{arXiv preprint arXiv:1409.1556} .

\bibitem[{Szegedy et~al.(2015)Szegedy, Liu, Jia, Sermanet, Reed, Anguelov,
  Erhan, Vanhoucke, and Rabinovich}]{szegedy2015going}
Szegedy, C.; Liu, W.; Jia, Y.; Sermanet, P.; Reed, S.; Anguelov, D.; Erhan, D.;
  Vanhoucke, V.; and Rabinovich, A. 2015.
\newblock Going deeper with convolutions.
\newblock In \emph{Proceedings of the IEEE conference on computer vision and
  pattern recognition}, 1--9.

\bibitem[{Tan and Le(2019)}]{tan2019efficientnet}
Tan, M.; and Le, Q.~V. 2019.
\newblock Efficientnet: Rethinking model scaling for convolutional neural
  networks.
\newblock \emph{arXiv preprint arXiv:1905.11946} .

\bibitem[{Tavanaei et~al.(2019)Tavanaei, Ghodrati, Kheradpisheh, Masquelier,
  and Maida}]{tavanaei2019deep}
Tavanaei, A.; Ghodrati, M.; Kheradpisheh, S.~R.; Masquelier, T.; and Maida, A.
  2019.
\newblock Deep learning in spiking neural networks.
\newblock \emph{Neural Networks} 111: 47--63.

\bibitem[{Taylor and Galton(2010)}]{taylor2010mostly}
Taylor, G.; and Galton, I. 2010.
\newblock A mostly-digital variable-rate continuous-time delta-sigma modulator
  ADC.
\newblock \emph{IEEE Journal of Solid-State Circuits} 45(12): 2634--2646.

\bibitem[{Wijekoon and Dudek(2009)}]{wijekoon2009cmos}
Wijekoon, J.~H.; and Dudek, P. 2009.
\newblock {A CMOS circuit implementation of a spiking neuron with bursting and
  adaptation on a biological timescale}.
\newblock In \emph{IEEE Biomedical Circuits and Systems Conference}, 193--196.

\bibitem[{Wu et~al.(2019{\natexlab{a}})Wu, Chua, Zhang, Yang, Li, and
  Li}]{wu2019deep}
Wu, J.; Chua, Y.; Zhang, M.; Yang, Q.; Li, G.; and Li, H. 2019{\natexlab{a}}.
\newblock Deep spiking neural network with spike count based learning rule.
\newblock In \emph{2019 International Joint Conference on Neural Networks
  (IJCNN)}, 1--6. IEEE.

\bibitem[{Wu et~al.(2015)Wu, Saxena, Zhu, and Balagopal}]{wu2015cmos}
Wu, X.; Saxena, V.; Zhu, K.; and Balagopal, S. 2015.
\newblock {A CMOS spiking neuron for brain-inspired neural networks with
  resistive synapses and in-situ learning}.
\newblock \emph{IEEE Transactions on Circuits and Systems II: Express Briefs}
  62(11): 1088--1092.

\bibitem[{Wu et~al.(2019{\natexlab{b}})Wu, Deng, Li, Zhu, Xie, and
  Shi}]{wu2019direct}
Wu, Y.; Deng, L.; Li, G.; Zhu, J.; Xie, Y.; and Shi, L. 2019{\natexlab{b}}.
\newblock Direct training for spiking neural networks: Faster, larger, better.
\newblock In \emph{Proceedings of the AAAI Conference on Artificial
  Intelligence}, volume~33, 1311--1318.

\bibitem[{Zhang et~al.(2018)Zhang, Yang, Ye, and Hua}]{zhang2018lq}
Zhang, D.; Yang, J.; Ye, D.; and Hua, G. 2018.
\newblock Lq-nets: Learned quantization for highly accurate and compact deep
  neural networks.
\newblock In \emph{Proceedings of the European conference on computer vision
  (ECCV)}, 365--382.

\bibitem[{Zhang et~al.(2019)Zhang, Zhou, Zhi, Du, and Chen}]{zhang2019tdsnn}
Zhang, L.; Zhou, S.; Zhi, T.; Du, Z.; and Chen, Y. 2019.
\newblock Tdsnn: From deep neural networks to deep spike neural networks with
  temporal-coding.
\newblock In \emph{Proceedings of the AAAI Conference on Artificial
  Intelligence}, volume~33, 1319--1326.

\bibitem[{Zhou et~al.(2020)Zhou, Chen, Li, and Sanyal}]{zhou2020deep}
Zhou, S.; Chen, Y.; Li, X.; and Sanyal, A. 2020.
\newblock Deep SCNN-based Real-time Object Detection for Self-driving Vehicles
  Using LiDAR Temporal Data.
\newblock \emph{IEEE Access} .

\bibitem[{Zhou and Li(2020)}]{zhou2020spiking}
Zhou, S.; and Li, X. 2020.
\newblock Spiking Neural Networks with Single-Spike Temporal-Coded Neurons for
  Network Intrusion Detection.
\newblock \emph{arXiv preprint arXiv:2010.07803} .

\end{thebibliography}

\newpage
\centerline{\Large{\bf Technical Appendix}}

\appendix

\section{Derivation of Spiking Neuron's Layer Response Models of Section \ref{subsection31}} \label{secA.3}

\subsection{General Solution to Membrane Potential}
Consider the integrate-and-fire spiking neuron model shown in Fig. \ref{fig:neuronfigure-append} with the membrane potential equation (\ref{eq2.10}). To solve for the membrane potential $v_j(t)$, one of the ways is to multiply $e^{bt}$ to both sides of (\ref{eq2.10})
\begin{equation}
      e^{bt}\frac{dv_j(t)}{dt}+be^{bt}v_j(t) = \sum_i w_{ji} \sum_k g(t-t_{ik})e^{bt},
\end{equation}
which can be re-written as
\begin{equation}
   \frac{d}{dt} \left( e^{bt}v_j(t) \right) = \sum_i w_{ji} \sum_k g(t-t_{ik}) e^{bt}. 
\end{equation}
Integrating both sides, we get
\begin{equation}
   \int_{0}^t \frac{d}{dx} \left( e^{bx}v_j(x) \right)dx = \sum_i w_{ji} \sum_k \int_{0}^t g(x-t_{ik}) e^{bx}dx, 
\end{equation}
which gives membrane potential
\begin{equation}
    v_j(t)=v_j(0)e^{-bt} + \sum_{i}w_{ji} e^{-bt} \sum_k \int_0^t g(x-t_{ik})e^{bx}dx.
\end{equation}
Assuming zero-initial condition $v_j(0)=0$ and causal spike waveform, i.e., $g(t)=0$ for $t<0$, we have the membrane potential expression
\begin{equation}
    v_j(t)=\sum_{i}w_{ji} e^{-bt} \sum_k \int_{t_{ik}}^t g(x-t_{ik})e^{bx}dx. \label{eq5.20}
\end{equation}
If $b>0$, then (\ref{eq5.20}) is for LIF neuron. For IF neuron, since $b=0$, the membrane potential becomes
\begin{equation}
    v_j(t)=\sum_{i}w_{ji} \sum_k \int_{t_{ik}}^t g(x-t_{ik})dx. \label{eq5.25}
\end{equation}
The neuron emits an output spike at time $t_{jk}$ whenever $v_j(t_{jk}) \geq \theta$.

We consider rate coding and temporal coding in this paper. For rate coding,
if the membrane potential is reset to zero after each spike, we can omit the spiking/resetting procedure, calculate the accumulative potential $v_j(T)$, and define the spike rate as 
\begin{equation}
    r_j = {\rm ReLU}\left(\frac{v_j(T)}{\theta T} \right).   \label{eq5.30}
\end{equation}
The function ${\rm ReLU}(x) = \max(0, x)$ is used to guarantee non-negative rate. 

Unfortunately, (\ref{eq5.30}) is valid for IF neurons only. It is not valid for LIF neurons because the effect of leaky is different between the reset membrane potential and the non-reset membrane potential. Jin et al. \cite{jin2018hybrid} applied (\ref{eq5.30}) to derive gradient for LIF neurons, which is not an accurate approach.  

As an alternative, assuming the spikes are regularly distributed in time, we can use the first spike's time $t_{j0}$ to calculate the spiking rate as
\begin{equation}
    r_j = \frac{1}{t_{j0}}.  \label{eq5.35}
\end{equation}

For temporal coding, we consider single-spike and time-to-first-spike (TTFS) coding, which means the first arrived spike among a set of spikes is the most significant. Each neuron emits only one spike and then resets to zero membrane potential for the rest of the time until $T$. In (\ref{eq5.20}) and (\ref{eq5.25}), we have $k=0$ only, so the second $\sum$ can be skipped. The input and output spike times can be simplified to $t_i = t_{i0}$ and $t_j = t_{j0}$, respectively.

For the spike waveform, we consider mainly the following three types: impulse waveform
\begin{equation}
    g(t) = \delta (t),   \label{eq5.10}
\end{equation}
Heaviside (unit-step) waveform
\begin{equation} 
  g(t) = a u(t) = \left\{ \begin{array}{ll} a, & t \geq 0 \\ 0, & {\rm else} \end{array} \right. \label{eq5.11}
\end{equation}
with constant $a$, and exponentially-decaying waveform
\begin{equation}
    g(t) = \left\{ \begin{array}{ll} \frac{1}{\tau} e^{-\frac{t}{\tau}}, & t \geq 0 \\ 0, & {\rm else}  \end{array} \right.  \label{eq5.12}
\end{equation}
We also call (\ref{eq5.12}) simply as the exponential waveform. Note that the exponentially-increasing waveform $e^{t/\tau}$ for $0\leq t \leq T$ will lead to similar analytic expressions. Note also that $\lim_{\tau \rightarrow 0} g(t) = \delta(t)$ and $\lim_{\tau \rightarrow \infty} a\tau g(t) =a u(t)$ for (\ref{eq5.12}). For rate coding, spike duration is limited by $1/r_i$. For temporal coding, spike duration is limited by $t_j-t_i \leq T$.

The above three spike waveforms are applied widely in SNN publications. The impulse and Heaviside waveform can be implemented easily in IBM TrueNorth digital neurons, while the exponential waveform is used in Intel Loihi. 
These three waveforms are perhaps the simplest ones that can lead to closed-form layer responses. Other more complex waveforms are usually lack of such closed-form solutions.

\subsection{Layer Response of Rate-Coded Neuron}

\paragraph{IF neuron with impulse spike:} In this case, $\sum_k \int_{t_{ik}}^T g(x-t_{ik})dx = Tr_i$, i.e., the total number of spikes in this time period. With rate definition (\ref{eq5.30}), from (\ref{eq5.25})  we can easily get layer response as 
\begin{equation}
    r_j = {\rm ReLU}\left(\sum_i \frac{w_{ji}}{\theta}  r_i \right).  \label{eq5.40}
\end{equation}
If using the rate definition (\ref{eq5.35}), we first calculate the first spike time $t_{j0}$ from
\begin{equation}
    v_j (t_{j0}) = \sum_i w_{ji} t_{j0} r_i  = \theta. 
\end{equation}
Then, we can get (\ref{eq5.40}) again based on (\ref{eq5.35}) and $t_{j0}$. It is interesting to observe that (\ref{eq5.40}) is almost identical to the DNN layer response, which means SNNs can be trained almost the same way as DNNs. Unfortunately, (\ref{eq5.40}) is derived under idealized assumptions such as $Tr_i$ or $t_{j0} r_i$ is integer, without which no closed-form expressions are available. Besides, it is assumed there is no potential loss during spike generation, which is not the case as shown in \cite{rueckauer2017conversion}. All these contribute to modeling errors that will build up and degrade the performance of deep SNNs.  

\paragraph{IF neuron with Heaviside spike:} Since the spike duration must be less than $1/r_i$ for each input neuron $i$, we assume $g(t)=a$ for $0 \leq t \leq \Delta T_i\leq 1/r_i$. To derive an analytic layer response, we need to assume that the input spikes are evenly distributed, which means the input spike time is $t_{ik}=k/r_i$ for $k=0, \cdots, Tr_i-1$. From (\ref{eq5.25}) we can obtain
\begin{equation}
    v_j(T) = \sum_{i} w_{ji} \sum_{k=0}^{Tr_i-1} \int_{0}^{\Delta T_i} g(x) dx = \sum_i w_{ji}a\Delta T_i T r_i.
\end{equation}
Using the rate definition (\ref{eq5.30}), we have
\begin{equation}
    r_j = {\rm ReLU} \left( \sum_i \frac{w_{ji}a\Delta T_i}{\theta} r_i \right), \label{eq5.45}
\end{equation}
which is similar to (\ref{eq5.40}). Obviously, spike length $\Delta T_i$ can not be equal to $1/r_i$. If using the rate definition (\ref{eq5.35}), we can find $t_{j0}$ by letting $v_j(t_{j0})=\theta$ and get the same expression as (\ref{eq5.45}). Note that \cite{hunsberger2016training} tried to derive $r_j$ expressions for a single input neuron using this similar way for training, but their results were different than ours and were questionable. 

\paragraph{IF neuron with exponential spike:} The spike waveform is time-limited as $g(t) = 1/\tau e^{-t/\tau}$ for $0 \leq t \leq 1/r_i$. From (\ref{eq5.25}) we can obtain
\begin{align}
    v_j(T) &= \sum_i w_{ji} \sum_{k=0}^{Tr_i-1} \int_{k/r_i}^{(k+1)/r_i} g(x-k/r_i) dx  \nonumber \\
    &= \sum_i w_{ji} T r_i(1-e^{-\frac{1}{\tau r_i}})
\end{align}
With rate definition (\ref{eq5.30}) we can find layer response as
\begin{equation}
    r_j = {\rm ReLU}\left( \sum_{i} \frac{w_{ji}}{\theta} r_i \left( 1-e^{-\frac{1}{\tau r_i}} \right)  \right). \label{eq5.50}
\end{equation}
When $\tau$ is small enough such that $e^{-1/\tau r_i} \rightarrow 0$, then (\ref{eq5.50}) becomes (\ref{eq5.40}). 
In addition, if using the rate definition (\ref{eq5.35}), we can find $t_{j0}$ by letting $v_j(t_j0)=\theta$ and get the same expression as (\ref{eq5.50}). 

\paragraph{LIF neuron with impulse spike:} For LIF neurons, if $g(t)=\delta(t)$, then from (\ref{eq5.20}), we have
\begin{align}
    v_j(t) &= \sum_i w_{ji} e^{-bt} \sum_{k=0}^{tr_i-1} e^{bk/r_i}  \nonumber \\
     & = \sum_i w_{ji} (1-e^{-bt})/(e^{b/r_i}-1).
\end{align}
Letting $v_j(t_{j0})=\theta$, we can find $t_{j0}$ and use rate definition (\ref{eq5.35}) to derive
\begin{equation}
    r_j = {\rm ReLU} \left(-b \log^{-1} \left(1-\frac{\theta}{\sum_i w_{ji}(e^{b/r_i}-1)^{-1}} \right) \right).   \label{eq5.60}
\end{equation}
Directly using (\ref{eq5.60}) in training does not work well because random weights $w_{ji}$ often make the $\log$ function undefined. As an alternative, we can convert (\ref{eq5.60}) to
\begin{equation}
    e^{-\frac{b}{r_j}} = 1-\frac{\theta}{\sum_i w_{ji}\frac{1}{1/e^{-b/r_i}-1}}, \label{eq5.62}
\end{equation}
and use $e^{-b/r_i}$ and $e^{-b/r_j}$ as neuron input and output in training. Unfortunately, this does not work well either because (\ref{eq5.62}) is extremely sensitive to $w_{ji}$ and $e^{-b/r_i}$. The left-hand-side of (\ref{eq5.62}) often becomes negative or saturated. The former will stop training, while the latter will make training stuck.

\paragraph{LIF neuron with Heaviside spike:} Consider $g(t) = a$ for $0 \leq t \leq \Delta T_i \leq 1/r_i$. Assume evenly distributed input spike times. From (\ref{eq5.20}), we can find
\begin{align}
    v_j(t) & = \sum_i w_{ji} e^{-bt} \sum_{k=0}^{tr_i-1} \int_{k/r_i}^{k/r_i+\Delta T_i} a e^{bx} dx \nonumber \\
    & = \sum_{i} w_{ji} \frac{a}{b}(e^{b\Delta T_i}-1) \frac{1-e^{-bt}} {e^{b/r_i}-1}
\end{align}
From $v_j(t_{j0})=\theta$, we can find $t_{j0}$. The layer response is thus
\begin{equation}
    r_j = {\rm ReLU}\left(-b \log^{-1}\left(1-\frac{\theta b/a(e^{b\Delta T_i}-1)^{-1}}{\sum_{i} w_{ji}(e^{b/r_i}-1)^{-1}} \right) \right). \label{eq5.65}
\end{equation}
We have the same problems in training as (\ref{eq5.60}) and (\ref{eq5.62}). 

\paragraph{LIF neuron with exponential spike:} For exponential spike waveform (\ref{eq5.12}), from (\ref{eq5.20}) we have
\begin{align}
    v_j(t) &= \sum_i w_{ji} e^{-bt} \sum_{k=0}^{tr_i-1} \int_{k/r_i}^{(k+1)/r_i} g(x-k/r_i)e^{bx} dx \nonumber \\
    & = \sum_i w_{ji} e^{-bt} \sum_{k=0}^{tr_i-1} \int_{0}^{1/r_i} \frac{1}{\tau} e^{-x/\tau} e^{b(x+k/r_i)} dx \nonumber \\
    & = \sum_{i} w_{ji} e^{-bt}  \frac{1}{b\tau-1}(1-e^{\frac{b\tau-1}{\tau r_i}}) \sum_{k=0}^{tr_i-1} e^{bk/r_i} \nonumber \\
    & = \frac{1-e^{-bt}}{b\tau-1} \sum_{i} w_{ji} (1-e^{\frac{b\tau-1}{\tau r_i}})\frac{1}{e^{b/r_i}-1}
\end{align}
From $v_j(t_{j0})=\theta$, we can find $t_{j0}$ and the layer response as
\begin{equation}
    r_j = {\rm ReLU} \left( -b \log^{-1} \left( 1- \frac{\theta (b\tau-1)}{\sum_i w_{ji} \frac{1-e^{(b\tau-1)/(\tau r_i)}} {e^{b/r_i}-1}} \right) \right).  \label{eq5.68}
\end{equation}
Obviously, We have the same problems in training as (\ref{eq5.60}) and (\ref{eq5.62}). 
    
Note that we have assumed $Tr_i$ and $t r_i$ be integers for all rate-based expressions. Such assumptions introduce modeling errors. Without such simplifying assumptions, no analytical expressions are available.

\subsection{Layer Response of Temporal-Coded Neuron}

\begin{figure}[t]
\centering
\fbox{\includegraphics[width=0.45\linewidth]{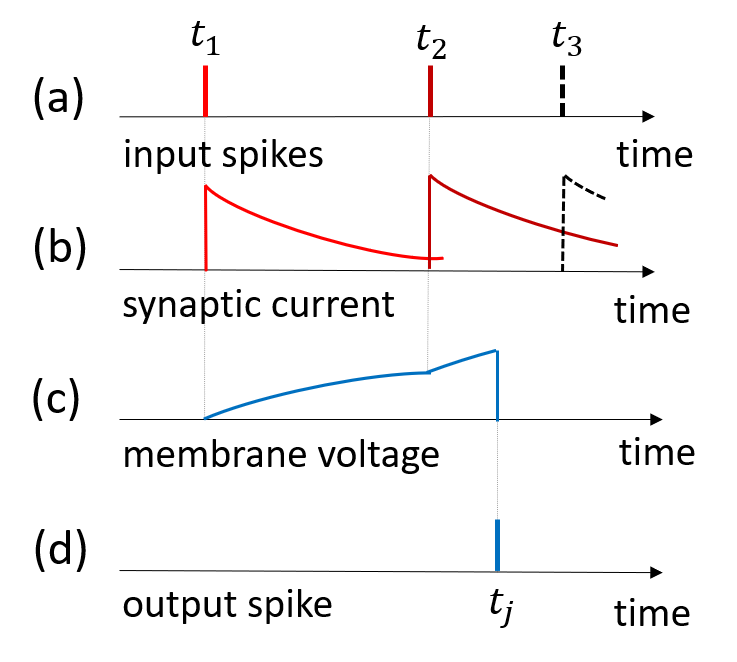}} \fbox{\includegraphics[width=0.45\linewidth]{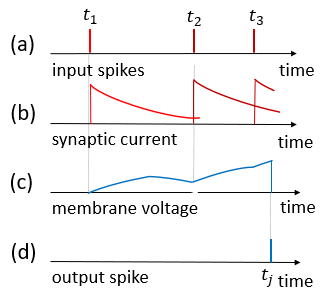}}
\centerline{(i) IF neuron\hspace{0.25\linewidth} (ii) LIF neuron}
\caption{Spiking time and waveform for (i) the IF neuron and (ii) the LIF neuron. (a) Three input neurons spike at time $t_1, t_2, t_3$, respectively. (b) Exponential synaptic current $g (t-t_k)$ jumps at time $t_k$ and extends through $T$. (c) Membrane potential $v_j(t)$ rises towards the firing threshold. (d) The output neuron $j$ emits a spike at time $t_j$ when the threshold is crossed.  The neurons support asynchronous processing and have neuron sparsity. As seen from Figure (i), the 3rd neuron will not spike since the output neuron spikes before it.}
\label{fig:neuroncircuit-append}
\end{figure}

For the integrate-and-fire neuron with single-spike temporal coding, a neuron is allowed to spike only once unless the network is reset or a new input pattern is presented. Fig. \ref{fig:neuroncircuit-append} illustrates how this neuron works for both the IF neuron and the LIF neuron. 

Impulse spikes are no longer appropriate because the neuron would have output spike time $t_j$ equal to $t_i$ for some input neuron $i$. This means that neuron spike time would become less and less diversified along with the increase of SNN layers. This will not lead to SNNs with good performance. Therefore, we consider only the Heaviside spike and the exponentially-decaying spike.

\paragraph{IF neuron with Heaviside spike:} With the spike waveform (\ref{eq5.11}), from (\ref{eq5.25}) we can find
\begin{align}
    v_j(t_j) = \sum_{i\in {\cal C}} w_{ji} \int_{t_i}^{t_j} g(x-t_i) dx 
       = \sum_{i\in {\cal C}} w_{ji} a(t_j - t_i) 
\end{align}
where the set ${\cal C}=\{\forall k: t_k < t_j\}$ includes all the input neurons (and only these neurons) that have spike time $t_k$ less than the output spike time $t_j$. The output spike time can be derived from $v_j(t_j) = \theta$ as
\begin{equation}
    t_j = \frac{\theta/a + \sum_{i\in {\cal C}} w_{ji}t_i} {\sum_{i \in {\cal C}} w_{ji}}.
\end{equation}
The above equation can be written as the standard DNN's layer response expression form (weight plus bias) as
\begin{equation}
    t_j =\sum_{i\in {\cal C}} t_i \frac{w_{ji}} {\sum_{\ell \in {\cal C}} w_{j\ell}} + \frac{\theta/a} {\sum_{\ell \in {\cal C}} w_{j\ell}}.     \label{eq3.25}
\end{equation}
Note that we do not use the ${\rm ReLU}$ notation to guarantee $t_j \geq 0$ because $t_j\geq T$ instead of $t_j=0$ if there is no valid solution or if the set ${\cal C}$ is empty. During searching for the set ${\cal C}$, we guarantee realistic $t_j \geq 0$.
The composite weights $\frac{w_{ji}} {\sum_{\ell \in {\cal C}} w_{j\ell}}$ and bias $\frac{\theta/a} {\sum_{\ell \in {\cal C}} w_{j\ell}}$ provide the necessary nonlinear activation. 

\paragraph{IF neuron with exponential spike:} In this case, 
the membrane potential at spiking time $t_{j}$ is
\begin{align}
    v_j(t_j) &= \sum_{i \in {\cal C}} w_{ji} \int_{t_i}^{t_j} \frac{1}{\tau} e^{-(x-t_i)/\tau} dx \nonumber \\
    &= \sum_{i\in {\cal C}} w_{ji} \left( 1- e^{-\frac{t_{j}-t_i}{\tau}} \right).
\end{align}
Considering the spiking threshold $\theta$, the neuron $j$'s spike time satisfies
\begin{equation}
e^{\frac{t_{j}}{\tau}} = \sum_{i\in {\cal C}} e^{\frac{t_{i}}{\tau}} \frac{w_{ji}} {\sum_{\ell \in {\cal C}} w_{j\ell}-\theta}. \label{eq3.26}
\end{equation}
As shown in \cite{mostafa2017supervised}, in the software implementation of SNN, we can directly use $e^{t_i/\tau}$ and $e^{t_j/\tau}$ as neuron activation, which makes the layer response (\ref{eq3.26}) similar to DNN's layer response. There is no bias term, but we can add one with $t_0 = 0$. We do not need other nonlinear activation because the composite weights ${w_{ji}}/({\sum_{\ell \in {\cal C}} w_{j\ell}-\theta})$ are nonlinear.

\paragraph{LIF neuron with Heaviside spike:}
For LIF neurons, from (\ref{eq5.20}) we can derive
\begin{equation}
    v_j(t_j)=\sum_{i\in {\cal C}}w_{ji} e^{-bt_j} \int_0^{t_j} g(x-t_i)e^{bx}dx. \label{eq3.60}
\end{equation}
With Heaviside spike waveform, assuming $t_j < T$, from (\ref{eq3.60}), at $v_j(t_j)=\theta$ we can get
\begin{align}
\theta &=\sum_{i\in {\cal C}}w_{ji} e^{-bt_j} \int_{t_i}^{t_j} a e^{bx}dx \nonumber \\
    &= \sum_{i\in {\cal C}} w_{ji} \frac{a}{b}\left( 1-e^{b(t_i-t_j)} \right).
\end{align}
Rearranging the terms, we arrive at
\begin{equation}
    e^{bt_j} = \sum_{i\in {\cal C}} e^{bt_i} \frac{w_{ji}}{\sum_{\ell \in {\cal C}}w_{j\ell} - b\theta/a},   \label{eq3.65}
\end{equation}
which is identical to the nonleaky IF neuron with an exponentially-decaying spike case (\ref{eq3.26}). This means IF neuron with an exponentially-decaying spike is the same as LIF neuron with a Heaviside spike for temporal-coding. 

\paragraph{LIF neuron with exponential spike:}
In this case, from (\ref{eq5.20}) we have
\begin{align}
  v_j(t_j)&=\sum_{i\in {\cal C}} w_{ji}e^{-bt_j} \int_{t_i}^{t_j} \frac{1}{\tau} e^{-\frac{x-t_i}{\tau}} e^{bx} dx    \nonumber \\
  &= \sum_{i\in {\cal C}}w_{ji}\frac{1}{1-b\tau} \left( e^{-\frac{t_j-t_i}{\tau}}-e^{-b(t_j-t_i)} \right). \label{eq3.70}
\end{align}
As pointed out in \cite{goltz2019fast}, there are only two special parameter settings that we can find a closed-form solution to $t_j$ for $v_j(t_j)=\theta$. The first parameter setting is $b\tau=1$ while the second parameter setting is $b\tau=1/2$.

For the first parameter setting $b\tau=1$, from (\ref{eq3.70}) we can derive 
\begin{equation}
    \lim_{b\rightarrow \frac{1}{\tau}} v_j(t_j) = \sum_{i\in {\cal C}} \frac{w_{ji}}{\tau} (t_j-t_i)e^{-b(t_j-t_i)}.  \label{eq3.72}
\end{equation}
At $v_j(t_j)=\theta$, rearranging the terms of (\ref{eq3.72}) we obtain
\begin{equation}
    \theta \tau e^{bt_j}-t_j\sum_{i\in{\cal C}}w_{ji}e^{bt_i} + \sum_{i \in {\cal C}}w_{ji}t_ie^{bt_i} =0,
\end{equation}
whose solution can be expressed as the Lambert $W$ function as
\begin{align}
    t_j = & \frac{\sum_{i\in{\cal C}}w_{ji}t_i e^{bt_i}} {\sum_{i\in{\cal C}}w_{ji}e^{bt_i}}  \nonumber \\
    & -\frac{1}{b}W\left(-\frac{b\theta \tau}{\sum_{i\in{\cal C}}w_{ji}e^{bt_i}} e^{\frac{b\sum_{i \in {\cal C}}w_{ji}t_i e^{bt_i}} {\sum_{i\in {\cal C}}w_{ji}e^{bt_i}}}
     \right).  \label{eq3.74}
\end{align}

For the second parameter setting $b\tau=1/2$, from (\ref{eq3.70}) and $v_j(t_j)=\theta$ we can get
\begin{equation}
 \theta \left( e^{\frac{t_j}{2\tau}} \right)^2 +
 \sum_{i\in{\cal C}} w_{ji}e^{\frac{t_i}{2\tau}} \left( e^{\frac{t_j}{2\tau}} \right) -
\sum_{i\in{\cal C}} w_{ji}e^{\frac{t_i}{\tau}}=0.
\end{equation}
We can find the solution as
\begin{align}
e^{\frac{t_j}{2\tau}}&=-\frac{1}{\theta}\sum_{i\in {\cal C}}w_{ji}e^{\frac{t_i}{2\tau}} \nonumber \\
& \pm \frac{1}{\theta} \sqrt{  \left(\sum_{i\in {\cal C}}w_{ji}e^{\frac{t_i}{2\tau}}\right)^2 + 2\theta \sum_{i \in {\cal C}} w_{ji}\left( e^{\frac{t_i}{2\tau}} \right)^2 }
\label{eq3.75}
\end{align}
It can be seen that $e^{\frac{t_j}{2\tau}}$ is a function of $e^{\frac{t_i}{2\tau}}$. Therefore, we can use $e^{\frac{t_i}{2\tau}}$ and $e^{\frac{t_j}{2\tau}}$ as the input and output neuron values in the software implementation of SNN.

Nevertheless, both (\ref{eq3.74}) and (\ref{eq3.75}) are unstable in training. Random weights $w_{ji}$ often make their left-hand-side to have negative or complex (non-real) values, which prevents gradient updating and stops training.

\section{Extra Experiment Results} \label{appendixB}

\subsection{Extra Experiment Results of MNIST}

\begin{figure}[t]
  \centering
  \includegraphics[width=0.45\textwidth]{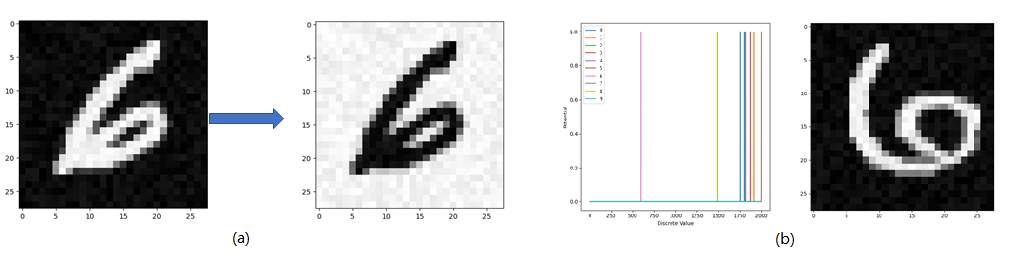}
  \caption{(a) Visualization of temporal coding from image pixel to spike time $t_i=\alpha(-p_i+1)$. (b) In the output layer, the $6$th neuron fired first, and thus the classification result was the digit 6.}
  \label{fig:mnistfigures}
\end{figure}

For MNIST, we trained our deep SNN model under two different scenarios: one with non-noisy input spike time and the other with noisy input spike time. As pointed out by \cite{mostafa2017supervised}, classification accuracy suffers if the temporal separation between the two consecutive output spiking times is decreased below the synaptic time constant. As described in Section \ref{sec4}, for the MNIST dataset, we applied the equation $t_i = \alpha(-p_i+1)$ to encode the input image pixel to spike time. We used $\alpha=3$ and normalized pixels $p_i \in [0, 1]$. Fig. \ref{fig:mnistfigures}(a) illustrates the original image and the encoded image (visualized by converting spike time to an image).
In the output layer, $10$ neurons were labeled from $0$ to $9$. In the example shown in Fig. \ref{fig:mnistfigures}(b), since the $6$th neuron fired first with the color ``pink", the image was recognized as $6$.

\begin{figure}[t]
  \centering{
  \includegraphics[width=0.45\textwidth]{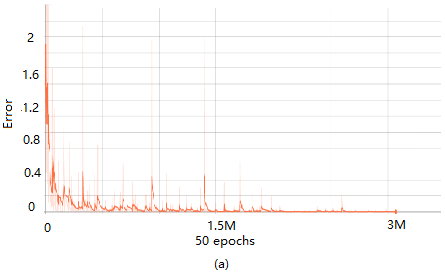}}
  \centering{
  \includegraphics[width=0.45\textwidth]{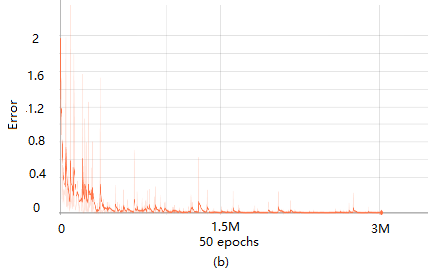}}
  \caption{Learning curves (loss/error versus epochs) during training the proposed SNN over the MNIST dataset: (a) non-noisy input case; (b) noisy input case. $50$ epochs were applied, with a total of 3 million mini-batch iterations.}
  \label{fig:mnisttraining}
\end{figure}

Regarding classification accuracy, the proposed model with noisy input had an accuracy of 99.33\%. The proposed model with non-noisy input had an accuracy of 98.82\%.
Fig. \ref{fig:mnisttraining} shows the learning curve, i.e., loss reduction as a function of training epochs, for the two models. The loss in both models converged to near 0 after several training epochs. After the first epoch, the accuracy reached 95.16\% for non-noisy input and 94.02\% for noisy input. After 10 epochs, the accuracy was 98.01\% for noisy input and 98.33\% for non-noisy input, which indicated high efficiency in training the SNNs.

\subsection{Extra Experiment Results of CIFAR-10}

We designed and experimented with three different deep SNN models for the CIFAR-10 dataset. The small one had 73K weights, the medium one had 11 million weights, and the large one (SpikingVGG16) had 33 million weights.
The temporal encoding rule was simply $t_i=\alpha p_i$ with $\alpha=3$, which ensured that the spike temporal separation was enlarged sufficiently to use as more pixels as needed. Besides, since the size of the CIFAR-10 dataset is not large, we exploited data augmentation to considerably increase the diversity of data available for training. Data augmentation was applied to the medium and large models, not the small model.

\begin{figure}[t]
  \centering{
  \includegraphics[width=0.4\textwidth]{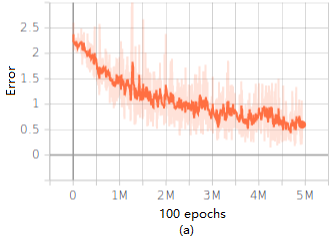}}
  \centering{
  \includegraphics[width=0.4\textwidth]{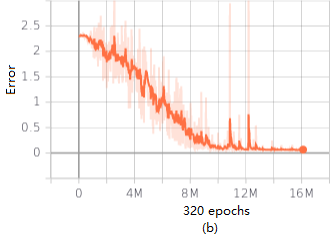}}
  \caption{(a) Learning curve for the small SNN model over CIFAR-10 without data augmentation.  (b) Learning curve for the medium SNN model over CIFAR-10 with data augmentation.}
  \label{fig:cifar10training}
\end{figure}

The small model consisted of three convolutional layers and two fully-connected layers: SCNN(5,64,2), SCNN(5,32,2), SCNN(5,16,2), FC(64), FC(10). This model was trained without data augmentation. The learning rate was 0.001 at the first epoch and was gradually decayed to 0.00001 at the 100th epoch. The learning curve of training this model can be observed from Fig. \ref{fig:cifar10training}(a). We applied 100 epochs or 5 million mini-batch iterations. The loss tended to level off after 75 epochs. The recognition accuracy was around 60\%. Note that the accuracy did not improve too much after the $20$th epoch. At the $20$th epoch, we had a recognition accuracy of 62.24\%, which was very high among similarly small networks.

The medium model consisted of four convolutional layers and three fully-connected layers: SCNN(3,64,1), SCNN(3,128,2), SCNN(3,256,1), SCNN(3,256,2), FC(1024), FC(1024), FC(10). This model was trained with data augmentation. The start learning rate was 0.001 and the end learning rate was 0.00001. We ran 320 epochs or 16 million mini-batch iterations during training. The learning curve was shown in Fig. \ref{fig:cifar10training}(b). After 240 epochs, the training tended to converge. The classification accuracy of this model was 80.49\%, close to the translation-based SNN of \cite{hunsberger2016training}, which had accuracy $82.95\%$. Nevertheless, our model had 11 million trainable weights, while the model of \cite{hunsberger2016training} had 39 million trainable weights which was much bigger. Data augmentation and bigger models can substantially increase the classification accuracy of SNNs.

\begin{figure}[t]
  \centering
  \includegraphics[width=0.48\textwidth]{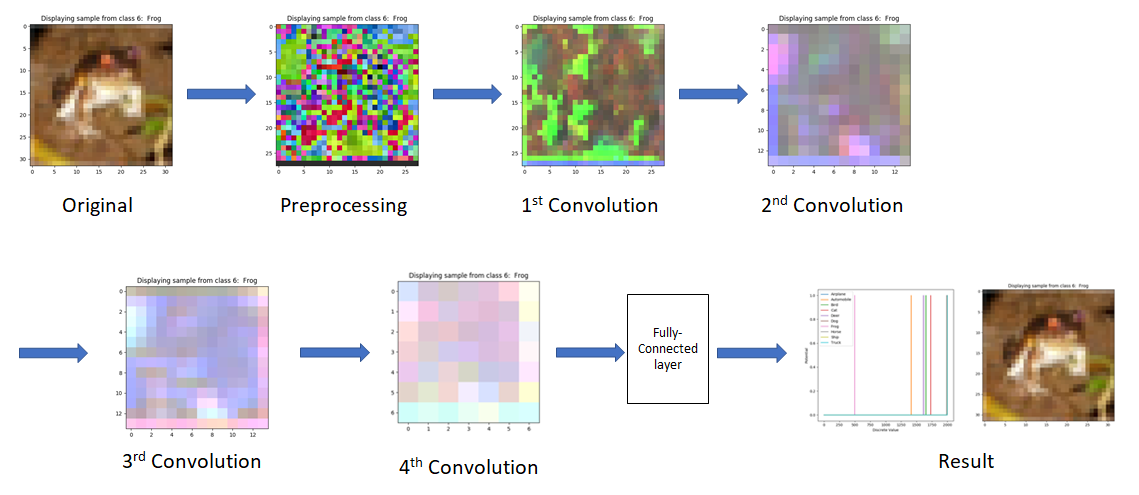}
  \caption{Visualization of the spiking activity of the proposed medium SNN model when classifying a CIFAR-10 sample image.}
  \label{fig:cifar10explain}
\end{figure}

Fig. \ref{fig:cifar10explain} provides a visualization of the medium model's neuron activation patterns. Preprocessing mapped the original pixel-based image to a spiking-time based image. At each SCNN layer, the neuron spiking activity was visualized as a feature map by converting neuron spiking time into pixel color. Dark-colored pixels mean the neurons had a large spike time or did not spike. At the output layer, ten neurons were labeled corresponding to the 10 object classes. In this figure, the output neuron labeled as ``frog" fired first, so the image was classified as ``frog".

Compared to the conventional DNNs, the classification accuracy of the above two models was still low. Hence, we built the large model SpikingVGG16 following the architecture of the DNN VGG16. Data augmentation was applied. Network architecture and experiment results were shown in Table \ref{tbl:modellist} and \ref{tbl:mnistacc}, respectively. In addition, since the energy consumption per neuron based on our proposed neuron circuit was $10$ pJ, the three proposed models consumed $731$ nJ, $0.112$ mJ and $0.336$ mJ, respectively.

\subsection{Extra Experiment Results of ImageNet}

\begin{figure}[t]
  \centering
  \includegraphics[width=0.45\textwidth]{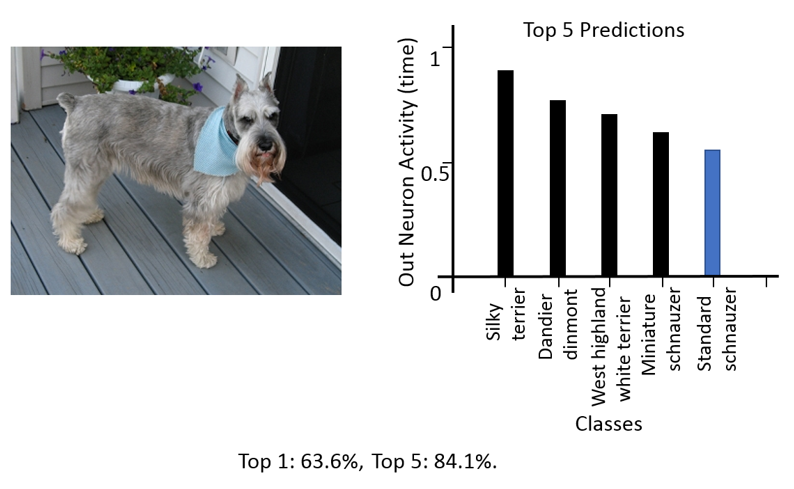}
  \caption{Classification results of the proposed SpikingGoogleNet over an ImageNet sample image.}
  \label{fig:imagenetexplain}
\end{figure}

\begin{figure*}[htbp]
  \centering
  \includegraphics[width=1.0\textwidth]{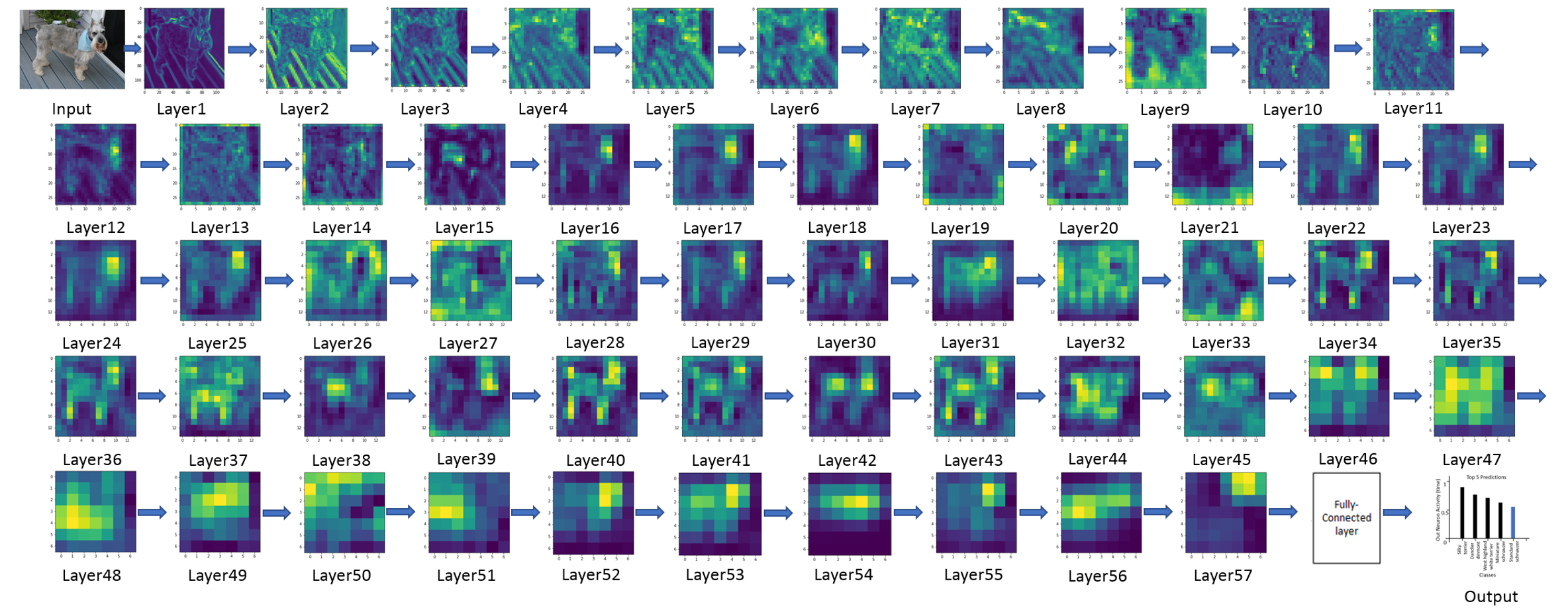}
  \caption{Visualization of the spiking activity of the SpikingGoogleNet when classifying an ImageNet sample image.}
  \label{fig:imagenetexplain3}
\end{figure*}

The architecture and the experiment results of our proposed SpikingGoogleNet were shown in Table \ref{tbl:modellist} and \ref{tbl:mnistacc}, respectively. Fig. \ref{fig:imagenetexplain} displayed the testing result of a randomly picked ImageNet image sample, whose Top-1 accuracy was 63.6\% and Top-5 accuracy was 84.1\%. The visualization of the neuron spiking activity of each layer is shown in Fig. \ref{fig:imagenetexplain3}. Note that there were plenty of dark-colored pixels, which indicated a high degree of sparsity.

\subsection{Extra Results of Weight Quantization}

\begin{figure*}[htbp]
  \centering
  \includegraphics[width=1.0\textwidth]{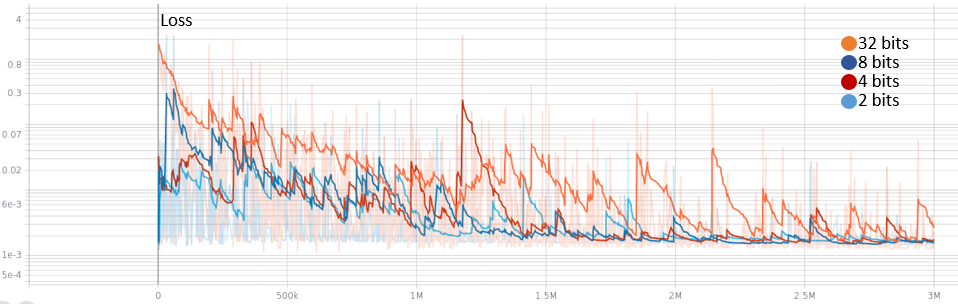}
  \caption{Learning curve of weight quantizing and retraining.}
  \label{fig:quantrain}
\end{figure*}

We conducted the weight quantization experiments with retraining to encode the weights in four different bit-widths, namely 32 bits, 8 bits, 4 bits, and 2 bits. Retraining was conducted in four stages. During the first stage, we trained the models with full-precision, i.e., 32-bit weights. After this stage converged, we quantized the weights to 8-bit and trained the models again until it converged. During the next stage, we quantized the 8-bit weights further to 4-bit and trained the models again until they converged. Similarly for the 2-bit weights.
Fig. \ref{fig:quantrain} shows how the training loss decreased during these four stages when the MNIST model was quantized and retrained. For each quantization level, we trained the model to converge to very low loss. When simply reducing the weights to a coarser quantization level, the loss was increased and retraining was conducted to make the model converge to low loss again.

The accuracy performance of the three extra SNN models with weight quantization that were not presented in Section \ref{sec4} is now listed in Table \ref{tbl:mnistaccquant2}. For the SNN model proposed for the MNIST dataset, whether with noisy input or non-noisy input, even though compressing the model size resulted in some loss of recognition accuracy, the largest loss was only about 0.18\% when using 2-bit quantization. Similarly, in CIFAR-10, for all the three SNN models, using 2-bit to store the weights dropped the accuracy no more than 1\%. Especially, from Table \ref{tbl:mnistaccquant}, for the large model with data augmentation, the recognition accuracy was still 90.93\%. Such results demonstrated
that our proposed SNN models were robust to weight quantization and
that in practical applications on hardware our networks could still perform very well.

\begin{table}[t]
  \centering
  \begin{tabular}{ccccc}
    \hline
    Model  &  \multicolumn{4}{c}{Weight Quantization (bits/weight)}      \\
                \cline{2-5}
         &   32  & 8  & 4  & 2  \\
    \hline
    MNIST non-noisy &  98.82 & 98.79 & 98.79 & 98.64 \\

    CIFAR-10 small  &  62.24 & 61.59 & 61.38 & 60.93 \\
    CIFAR-10 medium &  80.49 & 79.81  & 79.74 & 78.91 \\

    \hline
  \end{tabular}
    \caption{Classification Accuracy (\%) versus Weight Quantization.}
  \label{tbl:mnistaccquant2}
\end{table}


%
%

\subsection{Extra Results of Neuron Robustness to Random Hardware Mismatch}

\begin{figure}[htbp]
  \centering
  \includegraphics[width=0.4\textwidth]{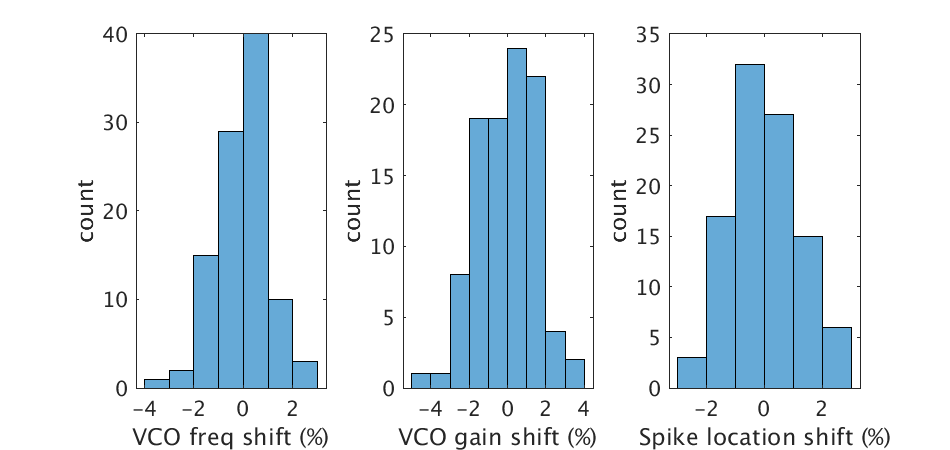}
  \caption{Monte-carlo random mismatch simulations showing distribution of (a) VCO center frequency, (b) VCO tuning gain, and (c) shift in output spike temporal position.}
  \label{fig:vcomismatch}
\end{figure}

Time-independent random mismatches in device geometry as well as variations in transistor threshold voltages are introduced into the design during the circuit fabrication step at the foundry which is beyond the control of designers. Threshold voltage variation will change center frequency and tuning gain of each VCO, and is problematic for a large neural network comprising of a multitude of neurons each consisting of 4 VCOs. We performed monte-carlo simulations on our neuron to evaluate the effect of random mismatches. Fig. \ref{fig:vcomismatch} (a) and (b) show variation in center frequency and tuning gain, respectively, of a VCO extracted from 100 monte-carlo runs using device mismatch models from the foundry. The VCO center frequency and tuning gain varied by $-4\% \sim +3\%$ and $\pm 4\%$ respectively. The shift in output spike location (timing jitter) due to random mismatch in the VCOs was in a similar range ($\pm 3\%$) as a shift in parameters for a single VCO as shown in Fig. \ref{fig:vcomismatch} (c). The smaller shift in output spike location is because the PLL forces VCO2 to track VCO1 even though VCO1's tuning gain and center frequency has shifted due to random mismatch, which eliminates the effect of mismatch in VCO2.

On the one hand, a $3\%$ timing jitter means an SNR of 30 dB, which shows that the neuron is robust to practical random hardware mismatch. On the other hand, our results also show that random mismatch contributes to a higher jitter in output spike than weight quantization or device noise, and thus may be a more significant limiting factor to SNN accuracy for the large neural network. The random mismatch can be reduced by increasing transistor sizes and/or using high threshold voltage transistors which will increase the area and reduce the speed of the SNN respectively. Thus, there is a trade-off between SNN accuracy, area, and speed due to random mismatches introduced during device fabrication.


\end{document}